%% file: main.tex
\def\isarxiv{1} %%% for icml submission version, we comment this line

\ifdefined\isarxiv
\documentclass[11pt]{article}

\usepackage[numbers]{natbib}

\else

\documentclass{article} 
\usepackage{microtype} 
\usepackage{graphicx}
\usepackage{subfig}
\usepackage{hyperref}
\usepackage{icml2024}

\fi

\usepackage{amsmath}
\usepackage{amsthm}
\usepackage{amssymb}
\usepackage{xcolor}
\usepackage{algpseudocode}
\ifdefined\isarxiv

\usepackage{algorithm}
\usepackage{grffile}%%% AAAI doesn't allow this
\usepackage{wrapfig,epsfig}
\usepackage{url}

\usepackage{epstopdf}

\usepackage{bbm} %%% AAAI doesn't allow this
\usepackage{dsfont}

\else

\fi 

\allowdisplaybreaks

\ifdefined\isarxiv

\usepackage{tikz}
\usepackage{hyperref}  %%% arxiv don't allow this.
\hypersetup{colorlinks=true,citecolor=blue,linkcolor=blue} %%% Zhao : maybe we should comment this in submission.
\usetikzlibrary{arrows}
\usepackage[margin=1in]{geometry}

\else

\fi
\theoremstyle{plain}
\newtheorem{theorem}{Theorem}[section]
\newtheorem{lemma}[theorem]{Lemma}
\newtheorem{definition}[theorem]{Definition}

\newtheorem{fact}[theorem]{Fact}
\newtheorem{remark}[theorem]{Remark}

\newcommand{\wt}{\widetilde}

\newcommand{\R}{\mathbb{R}}

\renewcommand{\tilde}{\wt}

\DeclareMathOperator*{\E}{{\mathbb{E}}}

\DeclareMathOperator{\cost}{cost}

\makeatletter
\newcommand*{\RN}[1]{\expandafter\@slowromancap\romannumeral #1@}
\makeatother
\newcommand{\Zhao}[1]{{\color{red}[Zhao: #1]}}

\usepackage{lineno}

\ifdefined\isarxiv

\else

\usepackage[textsize=tiny]{todonotes}

\icmltitlerunning{Dynamic Maintenance of Kernel Density Estimation Data Structure: From Practice to Theory}

\fi

\begin{document}

\ifdefined\isarxiv

\title{Dynamic Maintenance of Kernel Density Estimation Data Structure: From Practice to Theory}

\date{}

\author{
%\iffalse
Jiehao Liang\thanks{\texttt{jiehao.liang@berkeley.edu}. UC Berkeley.}
\and
Zhao Song\thanks{\texttt{zsong@adobe.com}. Adobe Research.}
\and 
Zhaozhuo Xu\thanks{\texttt{zx22@rice.edu}. Rice University.}
\and
Junze Yin\thanks{\texttt{junze@bu.edu}. Boston University.}
\and
Danyang Zhuo\thanks{\texttt{danyang@cs.duke.edu}. Duke University.}
%\fi
}

\else

\twocolumn[
\icmltitle{Dynamic Maintenance of Kernel Density Estimation Data Structure: From Practice to Theory}



\icmlsetsymbol{equal}{*}

\begin{icmlauthorlist}
\icmlauthor{Firstname1 Lastname1}{equal,yyy}
\icmlauthor{Firstname2 Lastname2}{equal,yyy,comp}
\icmlauthor{Firstname3 Lastname3}{comp}
\icmlauthor{Firstname4 Lastname4}{sch}
\icmlauthor{Firstname5 Lastname5}{yyy}
\icmlauthor{Firstname6 Lastname6}{sch,yyy,comp}
\icmlauthor{Firstname7 Lastname7}{comp}
\icmlauthor{Firstname8 Lastname8}{sch}
\icmlauthor{Firstname8 Lastname8}{yyy,comp}
\end{icmlauthorlist}

\icmlaffiliation{yyy}{Department of XXX, University of YYY, Location, Country}
\icmlaffiliation{comp}{Company Name, Location, Country}
\icmlaffiliation{sch}{School of ZZZ, Institute of WWW, Location, Country}

\icmlcorrespondingauthor{Firstname1 Lastname1}{first1.last1@xxx.edu}
\icmlcorrespondingauthor{Firstname2 Lastname2}{first2.last2@www.uk}

\icmlkeywords{Machine Learning, ICML}

\vskip 0.3in
]



\printAffiliationsAndNotice{\icmlEqualContribution} % otherwise use the standard text.

\fi

\ifdefined\isarxiv
\begin{titlepage}
  \maketitle
  \begin{abstract}
\input{abstract}

  \end{abstract}
  \thispagestyle{empty}
\end{titlepage}

%{\hypersetup{linkcolor=black}
%\tableofcontents
%}
%\newpage

\else
% \maketitle
\begin{abstract}
\input{abstract}
\end{abstract}

\fi

\input{intro} %%% Section 1. Introduction
\input{preli}

\input{technical}
\input{data}

%\input{correctness}
\input{adversary}

\input{conclusion}

% \newpage
\ifdefined\isarxiv
%\section*{Acknowledgments}
\bibliographystyle{alpha}
\bibliography{ref}
\else
\bibliography{ref}
\bibliographystyle{icml2024}
\fi

% \iffalse
\clearpage
\newpage
\onecolumn
\appendix
\section*{Appendix}
% {\hypersetup{linkcolor=black}
% \tableofcontents
% }

\input{app_preli}

\input{app_technical}

\input{app_data}

\input{correctness}

\input{app_correctness}

\input{app_adversary}

\input{app_lipschitz}
% \fi

%%%% Cut-line between first 10 pages and appendix

%%% some writing rules

%% Writing rule for creating tags.
%% Tags :
%% Theorem    \ref{thm:bla_bla}
%% Lemma      \ref{lem:bla_bla}
%% Claim      \ref{cla:bla_bla}
%% Corollary  \ref{cor:bla_bla}
%% Fact       \ref{fac:bla_bla}
%% Definition \ref{def:bla_bla}
%% Section    \ref{sec:bla_bla}
%% Subsection \ref{sub:bla_bla}
%% Equation   \ref{eq:bla_bla}

\end{document}

%% file: abstract.tex
Kernel density estimation (KDE) stands out as a challenging task in machine learning. The problem is defined in the following way: given a kernel function $f(x,y)$ and a set of points $\{x_1, x_2, \cdots, x_n \} \subset \mathbb{R}^d$, we would like to compute $\frac{1}{n}\sum_{i=1}^{n} f(x_i,y)$ for any query point $y \in \mathbb{R}^d$. Recently, there has been a growing trend of using data structures for efficient KDE. However, the proposed KDE data structures focus on static settings. The robustness of KDE data structures over dynamic changing data distributions is not addressed. In this work, we focus on the dynamic maintenance of KDE data structures with robustness to adversarial queries. Especially, we provide a theoretical framework of KDE data structures. In our framework, the KDE data structures only require subquadratic spaces. Moreover, our data structure supports the dynamic update of the dataset in sublinear time. Furthermore, we can perform adaptive queries with the potential adversary in sublinear time.

% \Zhao{To be clear, if we get 7pages+1 line, we desk rej. If we get 7pages - 1line, we also get desk rej.}\Jiehao{I'm adding more words in each Section.}

%% file: intro.tex
\section{Introduction}
Kernel density estimation~(KDE) is a well-known machine learning approach with wide applications in biology~\cite{fc17,crv+18}, physics~\cite{hallin2021classifying,c01} and law~\cite{chs20}. The KDE is defined as follows: given a kernel function $f(x,y)$ and a dataset $\{x_1, x_2, \cdots, x_n \}$, we would like to estimate 
\begin{align*}
\frac{1}{n}\sum_{i=1}^{n} f(x_i,y)
\end{align*}
for a query $y$. It is standard to assume the kernel $f$ to be positive semi-definite. From a statistics perspective, we regard KDE as estimating the density of a probability distribution provided by a mapping.

Recently, there has been a growing trend in applying probabilistic data structures for KDE  \cite{cs17,biw19,srb+19,acss20,ckns20,cs20,kap22}. The general idea is to transform the kernel function into a distance measure and then apply similarity search data structures such as locality sensitive hashing and sketching. This co-design of data structure and KDE is of practical importance: (1) the computational efficiency is taken into consideration when we design KDE algorithms; (2) the capacity of traditional probabilistic data structures is extended from search to sampling. As a result, we obtain a series of KDE algorithms with both sample efficiency and running time efficiency.

However, current KDE data structures focus on static settings where the dataset is fixed and the queries and independent of each other. More practical settings should be taken into consideration. In some applications of KDE, the dataset is dynamically changing. For instance, in time series modeling~\cite{mrl95,hl18}, the KDE estimators should be adaptive to the insertion and deletion in the dataset. In semi-supervised learning~\cite{whm+09,wlg19}, KDE data structures should handle the update of the kernel function. Moreover, in the works that apply KDE in optimization, the data structures should be robust over adversarial queries. As a result, the dynamic maintenance of KDE data structures should be emphasized in the research of machine learning.

In this paper, we argue that there exists a practice-to-theory gap for the dynamic maintenance of KDE data structures. Although there are existing work~\cite{ciu21} that supports insertion and deletion in KDE data structures, these operations' impact on the quality of KDE is not well-addressed. Moreover, the robustness of KDE data structures over adversaries is a recently raised concern. Thus, a formal theoretical analysis 
is required to discuss the robustness of KDE data structures in a dynamic setting.

In this paper, we present a theoretical analysis of the efficient maintenance of KDE for dynamic datasets and adversarial queries. Specifically, we present the first data structure design that can quickly adapt to updated input data and is robust to adversarial queries. We call our data structure and the corresponding algorithms \textit{adaptive kernel density estimation.}  
Our data structure only requires subquadratic spaces, and each update to the input data only requires sublinear time, and each query can finish in sublinear time.

\paragraph{Notations.}  

We use $\R$, $\R_+$, $\mathbb{N}_+$ to denote the set of real numbers, positive real numbers, and positive integers. For a set $X$, we use $|X|$ to denote its cardinality. Let $n \in \mathbb{N}_+$ and $r \in \R$. We define $[n] : = \{1, 2, 3, \dots, n\}$ and $\lceil r \rceil$ to be the smallest integer greater than or equal to $r$. $|r|$ is the absolute value of $r$. Let $\R^n$ be the set of all $n$-dimensional vectors whose entries are all real numbers. $\|x\|_2$ represents the $\ell_2$ norm of $x$. $\Pr[\cdot]$ represents the probability, and $\E[\cdot]$ represents the expectation. We define $\exp_2(r)$ as $2^r$.

\subsection{Related Work}

\paragraph{Efficient Kernel Density Estimation}
The naive KDE procedure takes a linear scan of the data points. This is prohibitively expensive for large-scale datasets. Therefore, it is of practical significance to develop efficient KDE algorithms. A series of traditional KDE algorithms, namely kernel merging~\cite{hs08,chm12}, is to perform clustering on the dataset so that the KDE is approximated by a weighted combination of centroids. However, these algorithms do not scale to high-dimensional datasets. 

On the other hand, there is a trend of sampling-based KDE algorithms. The focus of this line of work is to develop efficient procedures that approximate KDE with fewer data samples. Starting from random sampling~\cite{mf+17}, sampling procedures such as Herding~\cite{cws10} and $k$-centers~\cite{cs16} are introduced in KDE. Some work also incorporates sampling with the coreset concept~\cite{pt20} and provides a KDE algorithm by sampling on an optimized subset of data points. Recently, there has been a growing interest in applying hash-based estimators (HBE)~\cite{cs17,biw19,srb+19,cs20,cgc+18,ss21} for KDE. The HBE uses Locality Sensitive Hashing~(LSH) functions. The collision probability of two vectors in terms of an LSH function is monotonic to their distance measure. Using this feature, HBE performs efficient importance sampling by LSH functions and hash table type data structures. However, current HBEs are built for static settings and thus, are not robust to incremental changes in the input data. As a result, their application in large-scale online learning is limited. Except for LSH based KDE literature, there are also other KDE work based polynomial methods \cite{acss20,as23}. The dynamic type of KDE has also been considered in \cite{djs+22,bsz23}. The work \cite{dms23} presents both randomized algorithm deterministic algorithms for approximating a symmetric KDE computation.

\paragraph{Adaptive Data Structure}
Recently, there is a growing trend of applying data structures~\cite{clp+20,cmf+20,ssx21,xss21,xcl+21,sxz22,z22,sxyz22,wxw+22, lwz+23} to improve running time efficiency in machine learning. However, there exists a practice-to-theory gap between data structures and learning algorithms. Most data structures assume queries to be independent and provide theoretical guarantees based on this assumption. On the contrary, the query to data structures in each iteration of learning algorithms is mutually dependent. As a result, the existing analysis framework for data structures could not provide guarantees in optimization. To bridge this gap, quantization strategies~\cite{ssx21,xss21,sxz22,sxyz22} are developed for adaptive queries in machine learning. The idea of these approaches is to quantize each query into its nearest vertex on the $\epsilon$-net. Therefore, the failure probability of the data structures could be upper bounded by a standard $\epsilon$-net argument. Although quantization methods demonstrate their success in machine learning, this direct combination does not fully enable the power of both data structure and learning algorithms. In our work, we aim at a co-design of data structure and machine learning for efficiency improvements in adaptive KDE.

\subsection{Problem Formulation}

% \paragraph{Notations.}

In this work, we would like to study the following problem.

\begin{definition}[Dynamic Kernel Density Estimation]\label{def:dynamic_KDE}
Let $f:\mathbb{R}^d\times \mathbb{R}^d \rightarrow [0,1]$ denote a kernel function. Let   $X=\{x_{i}\}_{i=1}^{n} \subset \mathbb{R}^{d}$ denote a dataset. Let $$f_{\mathsf{KDE}}^{*}:=f(X,q):=\frac{1}{|X|}\sum_{x \in X} f(x,q)$$ define the kernel density estimate of a query $q\in \R^d$ with respect to $X$. Our goal is to design a data structure that efficiently supports any sequence of the following operations:  
\begin{itemize}
    \item \textsc{Initialize}$(f:\mathbb{R}^d\times \mathbb{R}^d \rightarrow [0,1],X\subset\mathbb{R}^d, \epsilon \in (0,1),f_{\mathsf{KDE}}\in[0,1])$. The data structure takes kernel function $f$, data set $X=\{x_1, x_2, \dots, x_n\}$, accuracy parameter $\epsilon$ and a known quantity $f_{\mathsf{KDE}}$ satisfying $f_{\mathsf{KDE}} \geq f_{\mathsf{KDE}}^{*}$ as input for initialization.
    \item \textsc{Update}$(z \in \R^d, i \in [n])$. Replace the $i$'th data point of data set $X$ with $z$.
    \item \textsc{Query}$(q \in \R^d)$. Output $\tilde{d}\in\mathbb{R}$  
    such that $$(1-\epsilon)f_{\mathsf{KDE}}^{*}(X,q)\leq \tilde{d} \leq(1+\epsilon)f_{\mathsf{KDE}}^{*}(X,q).$$
\end{itemize}
\end{definition}

We note that in the \textsc{Query} procedure do not assume i.i.d queries. Instead, we take adaptive queries and provide theoretical guarantees.

\subsection{Our Result}\label{sec:our_result}
In this work, we provide theoretical guarantees for the dynamic KDE data structures defined in Definition~\ref{def:dynamic_KDE}. We summarize our main result as below:

\begin{theorem}[Main result]\label{thm:main_result}
Given a function $K$ and a set of points set $X \subset \R^d$. Let $\cost(f)$ be defined as Definition~\ref{def:cost_K}. For any accuracy parameter $\epsilon \in (0,0.1)$, there is a data structure using space $O(\epsilon^{-2}n\cdot \cost(f))$  (Algorithm~\ref{alg:dynamic_KDE_initialize},~\ref{alg:dynamic_KDE_update} and~\ref{alg:dynamic_KDE_query}) for the Dynamic Kernel Density Estimation Problem (Definition~\ref{def:dynamic_KDE}) with the following procedures:
\begin{itemize}
    \item \textsc{Initialize}$(f:\mathbb{R}^d\times \mathbb{R}^d \rightarrow [0,1], X\subset\mathbb{R}^d, \epsilon \in (0,1),f_{\mathsf{KDE}}\in[0,0.1])$. Given a kernel function $f$, 
    a dataset $P$, an accuracy parameter $\epsilon$ and a quantity $f_{\mathsf{KDE}}$ as input, the data structure \textsc{DynamicKDE} preprocess in time
    \begin{align*}
        O(\epsilon^{-2}n^{1+o(1)}\cost(f)\cdot (\frac{1}{f_{\mathsf{KDE}}})^{o(1)}\log (1/ f_{\mathsf{KDE}}) \cdot \log^3 n)
    \end{align*}
    \item \textsc{Update}$(z \in \R^d, i \in [n])$. Given a new data point $z\in \mathbb{R}^d$ and index $i \in [n]$, the \textsc{Update} operation take $z$ and $i$ as input and update the data structure in time 
    \begin{align*}
        O(\epsilon^{-2}n^{o(1)}\cost(f)\cdot (\frac{1}{f_{\mathsf{KDE}}})^{o(1)}\log (1/ f_{\mathsf{KDE}}) \cdot \log^3 n)
    \end{align*}
    
    \item \textsc{Query}$(q \in \R^d)$. Given a query point $q\in \mathbb{R}^d$, the \textsc{Query} operation takes $q$ as input and approximately estimate kernel density at $q$ in time 
    \begin{align*}
        O(\epsilon^{-2}n^{o(1)}\cost(f)\cdot (\frac{1}{f_{\mathsf{KDE}}})^{o(1)}\log (1/ f_{\mathsf{KDE}}) \cdot \log^3 n).
    \end{align*}
    and output $\tilde{d}$ such that:
    \begin{align*}
        (1-\epsilon)f(X,q)\leq \tilde{d} \leq(1+\epsilon)f(X,q)
    \end{align*}
\end{itemize}
\end{theorem}
 
We will prove the main result in Lemma~\ref{lem:dynamic_KDE_initialize}, Lemma~\ref{lem:dynamic_KDE_update} and Lemma~\ref{lem:dynamic_KDE_query}.

\subsection{Technical Overview}
In this section, we introduce an overview of our technique that leads to our main result.

{\bf Density Constraint.}
We impose an upper bound on the true kernel density for query $q$. We also introduce geometric level sets so that the number of data points that fall into each level will be upper bounded.

{\bf Importance Sampling.}
To approximate kernel density efficiently, we adopt the importance sampling technique. We sample each data point with different probability according to their contribution to the estimation, the higher the contribution, the higher the probability to be sampled. Then, we can construct an unbiased estimator based on sampled points and sampling probability. The main problem is how to evaluate the contribution to KDE for each point. We explore the geometry property of the kernel function and estimate the contribution of each point based on their distance from the query point. 

{\bf Locality Sensitive Hashing.}
One problem with importance sampling is that when preprocessing, we have no access to the query point. It is impossible to estimate the contribution of each point for a future query. We make use of LSH to address this issue. To be specific, LSH preprocesses data points and finds the near neighbors for a query with high probability. With this property, we can first sample all data points in several rounds with geometrically decaying sampling probability. We design LSH for each round that satisfies the following property: given a query, LSH recovers points that have contributions proportional to the sampling probability in that round. Then we can find sampled points and proper sampling probability when a query comes.

{\bf Dynamic Update.}
Since the previous techniques, i.e. importance sampling and LSH, are independent of the coordinate value of the data point itself. This motivates us to support updating data points dynamically. Since LSH is a hash-based structure, given a new data point $z \in \R^d$ and index $i$ indicating the data point to be replaced, we search for a bucket where $x_i$ was hashed, replace it with new data point $z$ and update the hash table. Such an update will not affect the data structure for estimating kernel density.

{\bf Robustness.}
To make our data structure robust to adaptive queries, we take the following steps to obtain a robust data structure. Starting from the constant success probability, we first repeat the estimation several times and take the median. This will provide a high probability of obtaining a correct answer for a fixed point. Then we push this success probability to the net points of a unit ball. Finally, we generalize to all query points in $\R^d$. Thus we have a data structure that is robust to adversary query.

\paragraph{Roadmap} The rest of our paper is organized as follows: In Section~\ref{sec:preli}, we describe some basic definitions and lemmas that are frequently used. In Section~\ref{sec:technical}, we list some technical claims for our results. In Section~\ref{sec:data}, we demonstrate our data structure in detail, including the algorithm and the running time analysis. %We study the correctness of our data structure in Section~\ref{sec:correctness}. 
We perform an analysis of our data structures over the adversary in Section~\ref{sec:adversary}.
Finally, we draw our conclusion in Section~\ref{sec:conclusion}.

%% file: preli.tex
\section{Preliminary}\label{sec:preli}

The goal of this section is to introduce the basic definitions and lemmas that will be used to prove our main result. 

We first introduce a collection of subsets called geometric weight levels. 

\begin{definition}[Geometric Weight Levels]\label{def:level}
Fix $R \in\mathbb{N}_+$ and $q\in\R^d$. 
We define
%\begin{align*}
$
    w_i:=f(x_i,q)
$. 
%\end{align*}

For any fix $r\in[R]:=\{1,2,\cdots,R\}$, we define  
%$
\begin{align*}
    L_{r}:=\{x_{i}\in X ~|~ w_{i}\in(2^{-r+1},2^{-r}]\}.
\end{align*}
%$

We define the corresponding distance levels as
\begin{align*}
    z_r:=\max_{\mathrm{s.t.}f(z)\in(2^{-r},2^{-r+1}]} z.
\end{align*}
where $f(z):=f(x,q)$ for $z=\|x-q\|_2$. 

In addition, we define 
% \begin{align*}
$L_{R+1}:=X \setminus \bigcup_{ r \in [R] } L_{r}$.
% \end{align*}
\end{definition}

Geometric weight levels can be visualized as a sequence of circular rings centered at query $q$. The contribution of each level to kernel density at $q$ is mainly determined by the corresponding distance level.

Next, we introduce the importance sampling technique which is the key idea to accelerate the query procedure.

\begin{definition}[Importance Sampling]\label{def:importance_sampling}

Given data points $\{x_1, \cdots, x_n\} \subset \R^d$, we sample each point $x_i$ with probability $p_i$ and construct estimator as follows:
%\begin{align*}
$
    T:=\sum_{i=1}^{n} \frac{\chi_i}{p_i} x_i
$.
%\end{align*}

\end{definition}

To apply importance sampling, we need to evaluate the contribution of each point. We sample each point that has a high contribution with a high probability. 

A natural question arose: when preprocessing, we have no access to the query, so we cannot calculate distance directly. Locality Sensitive Hashing is a practical tool to address this problem. 

\begin{definition}[Locally Sensitive Hash \cite{im98}]\label{def:LSH_family}
A family $\mathcal{H}$ is called $(p_\mathrm{near},p_\mathrm{far},z,c)$-sensitive  where $p_\mathrm{near},p_\mathrm{far}\in [0,1],z \in\mathbb{R},c\geq 1$, if for any $x,q\in\mathbb{R}^d$:
\begin{itemize}
    \item $\Pr_{h\sim\mathcal{H}}[h(x)=h(q)~|~\| x-q\|_2\leq z]\geq p_\mathrm{near}$
    \item $\Pr_{h\sim\mathcal{H}}[h(x)=h(q)~|~\| x-q\|_2\geq cz]\leq p_{\mathrm{far}}$
\end{itemize}

\end{definition}

The next lemma shows the existence of the LSH family and its evaluation time.

\begin{lemma}[Lemma 3.2 in page 6  
of \cite{ai06}]\label{lem:p}
Let $(x, q) \in \mathbb{R}^d \times \mathbb{R}^d$. Define 
\begin{align*}
    p_\mathrm{near}:=p_1(z):=\Pr_{h\sim\mathcal{H}}[h(x)=h(q)~|~\| x-q\|_2 \leq z]
\end{align*}
and
\begin{align*}
    p_\mathrm{far}:=p_2(z,c):=\Pr_{h\sim\mathcal{H}}[h(x)=h(q)~|~\| x-q\|_2 \geq cz].
\end{align*}

% \Junze{The above 2 places: $\| x-q\|$ should be $\| x-q\|_2$}\Junze{Done.}

Then, if we fix $z$ to be positive,  we can have 
a hash family $\mathcal{H}$ satisfying
\begin{align*}
    \rho :=\frac{\log {1}/{p_\mathrm{near}}}{\log {1}/{p_\mathrm{far}}}\leq \frac{1}{c^2}+O(\frac{\log t}{t^\frac{1}{2}}),
\end{align*}
for any $c \geq 1, t > 0$, where $p_\mathrm{near}\geq e^{-O(\sqrt{t})}$ and it requires $dt^{O(t)}$ time for every evaluation.

\end{lemma}

\begin{remark}\label{rmk:p_near}
We set $t=\log^{\frac{2}{3}}n$, which results in $n^{o(1)}$ evaluation time and $\rho = \frac{1}{c^2}+o(1)$. Note that if $c=O(\log^\frac{1}{7}n)$, then 
\begin{align*}
    \frac{1}{\frac{1}{c^2}+O(\frac{\log t}{t^\frac{1}{2}})}=c^2(1-o(1)). \footnote{The above three $o(1)$ can be $\frac{\log \log^\frac{2}{3}n}{\log ^\frac{1}{3}n}, \frac{\log \log^\frac{2}{3}n}{\log ^\frac{1}{3}n}, \frac{\log \log^\frac{2}{3}n}{\log ^\frac{1}{21}n}$ respectively.}
\end{align*}

\end{remark}
 
Next, we assign the LSH family to each geometric weight level (Definition~\ref{def:level}) and show how well these families can distinguish points from different levels.

\begin{lemma}[probability bound for separating points in different level sets, informal version of Lemma~\ref{lem:LSH_formal}]\label{lem:LSH}

Given kernel function $f$ and $r \in [R]$, let $L_r$ be the weight level set and $z_r$ be the corresponding distance level (Definition~\ref{def:level}). For any query $q\in\R^d$, any integer pair  $(i, r) \in [R+1] \times [R]$, satisfying $i>r$, let $x \in L_r$ and $x' \in L_i $. Let  $$c_{i,r}:=\min\{\frac{z_{i-1}}{z_r},\log^{1/7}n\}.$$ We set up Andoni-Indyk LSH family (Definition~\ref{def:LSH_family}) $\mathcal{H}$ with near distance $z_r$ and define
\begin{align*}
    p_{\mathrm{near},r}&~:=\Pr_{h\sim\mathcal{H}}[h(x)=h(q)~|~\| x-q\|_2\leq z]
    % p_{\mathrm{far},r}&~:=\Pr_{h\sim\mathcal{H}}[h(x)=h(q)~|~\| x-q\|_2\geq cz]
\end{align*}
and
\begin{align*}
    % p_{\mathrm{near},r}&~:=\Pr_{h\sim\mathcal{H}}[h(x)=h(q)~|~\| x-q\|_2\leq z]
    p_{\mathrm{far},r}&~:=\Pr_{h\sim\mathcal{H}}[h(x)=h(q)~|~\| x-q\|_2\geq cz].
\end{align*}

Then, for any $k\geq 1$, it is sufficient to show the following:
\begin{enumerate}
    \item $\Pr_{h^*\sim \mathcal{H}^k}[h^*(x)=h^*(q)]\geq p_{\mathrm{near},r}^k$
    \item $\Pr_{h^*\sim \mathcal{H}^k}[h^*(x')=h^*(q)]\leq p_{\mathrm{near},r}^{kc_{i,r}^2(1-o(1))}$
\end{enumerate}
 
\end{lemma}

This lemma suggests that we can apply LSH several times to separate points in different level sets. It is useful for recovering points in a specific level set when estimating the ``importance" of a point based on its distance from the query point. We will discuss more in Section~\ref{sec:data}. 
We use a similar definition for the cost of the kernel in \cite{ckns20}.  
\begin{definition}[Kernel cost]\label{def:cost_K}
Given a kernel $f$, which has geometric weight levels $L_r$'s and distance levels $z_r$'s defined in Definition~\ref{def:level}. For any $r\in[R]$, we define the kernel cost $f$ for $L_r$ as  
\begin{align*}
    \cost(f,r):=\exp_2(\max\limits_{i\in\{r+1,\cdots,R+1\}}\lceil\frac{i-r}{c_{i,r}(1-\mathrm{o}(1))}\rceil),
\end{align*}
where 
% \begin{align*}
$c_{i,r}:=\min\{\frac{z_{i-1}}{z_{r}},\log^\frac{1}{7}n\}$.
% \end{align*}

Then we define the general cost of a kernel $f$ as
\begin{align*}
    \cost(f):=\max_{r \in [R]}\cost(f,r).
\end{align*}
\end{definition}
Note that when $f$ is Gaussian kernel, the $\cost(f)$ is $(\frac{1}{f_{\mathsf{KDE}}})^{(1+o(1))\frac{1}{4}}$ \cite{ckns20}.

%% file: technical.tex
\section{Technical Claims}\label{sec:technical}

In this section, we list some technical claims that are useful for our main results.

We start by giving an upper bound on sizes of geometric weight levels (Definition~\ref{def:level}).

\begin{lemma}[Sizes of geometric weight levels]\label{lem:upper_bound_geometric_informal}
Given $r\in[R]$, we have
\begin{align*}
    |L_r|\leq 2^r n f_{\mathsf{KDE}}^*\leq 2^r n f_{\mathsf{KDE}}.
\end{align*}
\end{lemma}

Next, we show a lemma for the probability of recovering a point in the query procedure, given that this point is sampled in the preprocessing stage.

\begin{lemma}[Probability for sampled point recovery]\label{lem:lower_bound_recovered_point_informal}
Suppose that we invoke \textsc{DynamicKDE.Initialize}. Suppose when $a=a^*$ and $r=r^*$, we sample a point $x \in L_{r^*}$. Given a query $q$, we invoke \textsc{DynamicKDE.Query}. With probability at least $1-\frac{1}{n^{10}}$, $H_{a^*,r^*}$ recovered $x$.
\end{lemma}

With the above lemma, we can bound the number of recovered points in expectation. We show that there are only $O(1)$ points recovered by LSH in each geometric weight level (Definition~\ref{def:level}).

\begin{lemma}[Upper bound on number of recovered points in expectation]\label{lem:upper_bound_recovered_point_informal}
Fix a query $q\in\mathbb{R}^d$. We define $R:=\lceil \log \frac{1}{f_{\mathsf{KDE}}} \rceil$. Fix $r \in [R]$, we define $ p:=p_{\mathrm{near},r}$. For each $(i,r)\in [R] \times [R]$,
% \Junze{I think $[R] \times [R]$ is better than $[R]^2$}\Junze{Done.},
we define $$c_{i,r}:=\min\{\frac{z_{i-1}}{z_r},\log^\frac{1}{7}n\}.$$  
There exists $k\in\mathbb{N}_+$ 
\begin{align*}
    k:=k_r:=\frac{1}{\log (1/p)}\max\limits_{l\in\{r+1,\cdots,R+1\}}\lceil\frac{l-r}{c_{l,r}^2(1-o(1))}\rceil,
\end{align*}
such that for any $i>j$
\begin{align*}
    \E_{h^*\sim \mathcal{H}^{k}}[|\{x^\prime \in L_i:h^*(x^\prime)=h^*(q)\}|]=O(1)
\end{align*}
\end{lemma}

Finally, we claim that the kernel function is Lipschitz. This is an important property for designing robust algorithms.

\begin{lemma}[Lipschitz property of KDE function]
Suppose kernel function $f^*_{\mathsf{KDE}}:\R^d\times\R^d\rightarrow[0,1]$ satisfies the following properties:
\begin{itemize}
    \item Radial: there exists a function $f:\R\rightarrow[0,1]$ such that $f(p,q)=f(\|p-q\|_2)$, for all $p,q\in\R$.
    \item Decreasing: $f$ is decreasing
    \item Lipschitz: $f$ is $\mathcal{L}$-Lipschitz
\end{itemize}
Then KDE function $f_{\mathsf{KDE}}^*:\R^d\rightarrow [0,1]$, 
\begin{align*}
f_{\mathsf{KDE}}^*(q):=\frac{1}{|P|}\sum_{p\in P}f(p,q)
\end{align*}
is $\mathcal{L}$-Lipschitz, i.e.
\begin{align*}
    |f_{\mathsf{KDE}}^*(q)-f_{\mathsf{KDE}}^*(q')|\leq \mathcal{L}\cdot\|q-q'\|_2
\end{align*}
\end{lemma}

%% file: data.tex
\section{Our Data Structures}\label{sec:data}

Our algorithm's high-level idea is the following. We apply importance sampling (Definition~\ref{def:importance_sampling}) to approximate kernel density at a query point $q$ efficiently. We want to sample data points with probability according to their contribution to the estimation. Ideally, given query point $q \in \R^d$ and data set $X \subset \R^d$, we can sample each data point in $X$ with probability proportional to the inverse of the distance from query $q$. Unfortunately, we have no access to query points when preprocessing. Hence we make use of LSH (Definition~\ref{def:LSH_family}) to overcome this problem.

In general, given a query $q$, LSH can recover its near points with high probability while the probability of recovering far points is bounded by a small quantity. To apply LSH, we first run 
\begin{align*}
R=\lceil \frac{1}{f_{\mathsf{KDE}}} \rceil
\end{align*}
rounds sampling, in which we sample each data point with probability $\frac{1}{2^r n f_{\mathsf{KDE}}}$ in $r$-th round. Then we obtain $R$ subsampled data sets. Given query $q$, we use LSH to recover those points both in level set $L_r$ and the $r$-th subsampled data sets. Hence we obtain the sampled data points and the corresponding sampling rates (in other words ``importance''). Then we construct the estimator as in Definition~\ref{def:importance_sampling}. 

Finally, we repeat the estimation process independently and take the median to obtain $(1 \pm \epsilon)$ approximation with high probability.

\subsection{LSH Data Structure}\label{app:data_structure_lsh_formal}

In this section, we present our LSH data structure with the following procedures:

{\bf Initialize.} Given a data set $\{x_1, \cdots, x_n\} \in \R^d$ and integral parameters $k, L$, it first invokes private procedure \textsc{ChooseHashFunc}. The idea behind this is to amplify the ``sensitivity'' of hashing by concatenating $k$ basic hashing functions from the family $\mathcal{H}$(Algorithm~\ref{alg:LSH_public_app} line~\ref{lin:basic_hash_family}) into new functions. Thus we obtain a family of ``augmented'' hash function $\mathcal{H}_l, l \in [L]$ (Algorithm~\ref{alg:LSH_private_app} line~\ref{lin:LSH_sample_k_functions}). We follow by \textsc{ConstructHashTable} in which we hash each point $x_i$ using the hashing function $\mathcal{H}_{l}$. Then we obtain $L$ hash tables corresponding to $L$ hash functions which can be updated quickly.

{\bf Recover.} Given a query $q \in \R^d$, it finds the bucket where $q$ is hashed by $\mathcal{H}_l$ and retrieves all the points in the bucket according to hashtable $\mathcal{T}_l$. This operation applies to all $L$ hashtables.

{\bf UpdateHashTable.} Given a new data point $z \in \R^d$ and index $i \in [n]$, it repeats the following operations for all $l \in [L]$: find bucket $\mathcal{H}_l(z)$ and insert point $z$; find bucket $\mathcal{H}_l(x_i)$ and delete point $x_i$.

Note that traditional LSH data structure only has \textsc{Initialize} and \textsc{Recover} procedures. To make it a dynamic structure, we exploit its hash storage. We design \textsc{UpdateHashTable} procedure so that we can update the hash table on the fly. This procedure provides guarantee for dynamic kernel density estimation.

\subsection{Initialize Part of Data Structure}

\begin{algorithm}[!ht]\caption{Dynamic KDE, members and initialize part, informal version of Algorithm~\ref{alg:dynamic_KDE_initialize}}\label{alg:dynamic_KDE_initialize_pseudo}
\begin{algorithmic}[1]
\State {\bf data structure} \textsc{DynamicKDE}      \Comment{Theorem~\ref{thm:main_result}}
\State
\State {\bf members} 
    \State \hspace{4mm} For ${i\in[n]}$,$x_i \in \mathbb{R}^d$ \Comment{dataset $X$}
    \State \hspace{4mm} $K_1, R, K_2 \in \mathbb{N}_+$ \Comment{Number of repetitions} 
    \State \hspace{4mm} For $a \in [K_1]$, $r \in [R]$, $H_{a,r}\in \textsc{LSH}$ \label{lin:instance_LSH} \Comment{Instances from \textsc{LSH} class}
\State {\bf end members}
\State
\Procedure{\textsc{Initialize}}{$X \subset \R^d,\epsilon \in (0,1),f_{\mathsf{KDE}} \in [0,1]$} \Comment{Lemma~\ref{lem:init}}
    \State Initialize $K_1,R$ as in Section~\ref{app:data}
    \For{$a=1,2,\cdots,K_1$}
        \For{$r=1,2,\cdots,R$}
            \State Compute $K_{2,r}, k_r$ as in Section~\ref{app:data}
            \State $P_j\leftarrow$ sample each element in $X$ with probability $\min\{\frac{1}{2^r n f_{\mathsf{KDE}}}, 1\}$.
            \State $\mathcal{H}_{a,r}.\textsc{Initialize}(P_r,k_r,K_{2,r})$ 
        \EndFor
        \State $\tilde{P}_{a}\leftarrow$ sample elements from $X$, each one has sample probability $\frac{1}{n}$ \Comment{Store $\tilde{P}_a$}
    \EndFor
\EndProcedure
\State
\State {\bf end data structure}
\end{algorithmic}
\end{algorithm}

In this section, we present the initialize part of our data structure.
We start by analyzing the space storage for \textsc{LSH} and \textsc{DynamicKDE}. Then we state the result of running time for \textsc{LSH} in Lemma~\ref{lem:LSH_initialize_upper_bound} and \textsc{DynamicKDE} in Lemma~\ref{lem:dynamic_KDE_initialize}. 
We first show the space storage of LSH part in our data structure.

\begin{lemma}[Space storage of \textsc{LSH}, informal version of Lemma~\ref{lem:LSH_storage_formal}]\label{lem:LSH_storage}
Given data set $\{x_i\}_{i\in[n]}\subset\R^d$, parameter $L,k\in\mathbb{N}_+$, the \textsc{Initialize} (Algorithm~\ref{alg:LSH_public_app}) of the data-structure \textsc{LSH} uses space
\begin{align*}
    O(Lkdn^{o(1)}+Ln)
\end{align*}
\end{lemma} 

Using the lemma above, we can prove the total space storage of \textsc{DynamicKDE} structure in the following lemma.

\begin{lemma}[Space storage part of Theorem~\ref{thm:main_result}, informal version of Lemma~\ref{lem:init_formal}]\label{lem:init}
The \textsc{Initialize} of the data structure \textsc{DynamicKDE} (Algorithm~\ref{alg:dynamic_KDE_initialize}) uses space
\begin{align*}
    & ~ O(\epsilon^{-2}(\frac{1}{f_{\mathsf{KDE}}})^{o(1)}\cdot\log(1 / f_{\mathsf{KDE}})\cdot \\
    & ~ \mathrm{cost}(K)  \cdot\log^2 n\cdot(\frac{1}{f_{\mathsf{KDE}}}+n^{o(1)}\cdot\log^2 n)) 
\end{align*}
\end{lemma}

For the running time, we again start with the LSH part.

\begin{lemma}[Upper bound on running time of \textsc{Initialize} of the data-structure \textsc{LSH}, informal version of Lemma~\ref{lem:LSH_initialize_upper_bound_formal} ]\label{lem:LSH_initialize_upper_bound}
Given input data points $\{x_i\}_{i\in[n]}\subset \mathbb{R}^d$, parameters $k,L\in \mathbb{N}_+$, LSH parameters $p_{\mathrm{near}},p_{\mathrm{far}} \in [0,1],c\in[1,\infty), r\in \R_+$ and kernel $K$, the \textsc{Initialize} of the data-structure \textsc{LSH}(Algorithm~\ref{alg:LSH_public_app}) runs in time
\begin{align*}
    O(L\cdot(kdn^{o(1)}+dn^{1+o(1)}+n\log n))
\end{align*}
\end{lemma}

Having shown the running time of LSH, we now move on to prove the total running time of \textsc{Init} in our data structure by combining the above result in the LSH part.

\begin{lemma}[The initialize part of Theorem~\ref{thm:main_result}, informal version of Lemma~\ref{lem:dynamic_KDE_initialize_formal}]\label{lem:dynamic_KDE_initialize}
Given $(K:\mathbb{R}^d\times \mathbb{R}^d \rightarrow [0,1], X \subset\mathbb{R}^d, \epsilon \in (0,1),f_{\mathsf{KDE}}\in[0,1])$, the \textsc{Initialize} of the data-structure \textsc{DynamicKDE} (Algorithm~\ref{alg:dynamic_KDE_initialize}) runs in time
\begin{align*}
    O(\epsilon^{-2}n^{1+o(1)}\cost(f)\cdot (\frac{1}{f_{\mathsf{KDE}}})^{o(1)}\log(1 / f_{\mathsf{KDE}}) \cdot \log^2 n)
\end{align*}
\end{lemma}

\begin{algorithm}[!ht]\caption{Dynamic KDE, update part, informal version of Algorithm~\ref{alg:dynamic_KDE_update}}\label{alg:dynamic_KDE_update_pseudo}
\begin{algorithmic}[1]
\State {\bf data structure} \textsc{DynamicKDE} \Comment{Theorem~\ref{thm:main_result}}
 \State
\Procedure{\textsc{Update}}{$v \in \R^d, f_{\mathsf{KDE}} \in [0,1], i \in [n]$} \Comment{Lemma~\ref{lem:dynamic_KDE_update}}
    \For{$a=1,2,\cdots,K_1$}
        \For{$r=1,2,\cdots,R$}
            \State Update the hashtables in $\mathcal{H}_{a,r}$ with $v$
        \EndFor
    \EndFor
    \State Replace $x_i$ with $v$
\EndProcedure
\State
\State {\bf end data structure}
\end{algorithmic}
\end{algorithm}

\subsection{Update Part of Data Structure}

In this section, we move to the update part of our data structure.
We first prove how to update the LSH data structure. Then we can extend the update procedure to \textsc{DynamicKDE} structure so that we can prove Lemma~\ref{lem:dynamic_KDE_update}.

\begin{lemma}[Update time of LSH, informal version of Lemma~\ref{lem:hash_update_formal}]\label{lem:hash_update}
Given a data point $v \in\mathbb{R}^d$ 
% \Junze{This should be $\mathbb{R}^d$}\Junze{DOne.} 
and index $i\in[n]$, the \textsc{UpdateHashTable} of the data-structure \textsc{LSH} (Algorithm~\ref{alg:LSH_public_app}) runs in (expected) time
\begin{align*}
    O(n^{o(1)}\log(n)\cdot \cost(f)).
\end{align*}
 
\end{lemma}

\textsc{Update} in LSH structure is a key part in \textsc{Update} for \textsc{DynamicKDE}.
Next, we show the running time of \textsc{Update} for \textsc{DynamicKDE} by combining the above results.

\begin{lemma}[The update part of Theorem~\ref{thm:main_result}, informal version of Lemma~\ref{lem:dynamic_KDE_update_formal}]\label{lem:dynamic_KDE_update}
Given an update $v \in\R^d$ and index $i\in[n]$, the \textsc{Update} of the data-structure \textsc{DynamicKDE} (Algorithm~\ref{alg:dynamic_KDE_update}) runs in (expected) time
\begin{align*}
    O(\epsilon^{-2}n^{o(1)}\cost(f)\cdot (\frac{1}{f_{\mathsf{KDE}}})^{o(1)}\log(1 / f_{\mathsf{KDE}}) \cdot \log^3 n).
\end{align*}
\end{lemma}

\subsection{Query Part of Data Structure}

Finally, we come to the query part. The goal of this section is to prove Lemma~\ref{lem:dynamic_KDE_query}, which shows the running time of \textsc{Query} procedure for \textsc{DynamicKDE}.

\begin{algorithm}[!ht]\caption{Dynamic KDE, query part, informal version of Algorithm~\ref{alg:dynamic_KDE_query}}\label{alg:dynamic_KDE_query_pseudo}
\begin{algorithmic}[1]
\State {\bf data structure} \textsc{DynamicKDE} \Comment{Theorem~\ref{thm:main_result}}
\State
 
\Procedure{\textsc{Query}}{$q\in \mathbb{R}^d, \epsilon \in (0,1),f_{\mathsf{KDE}} \in [0,1]$} 
 
    \For{$a=1,2,\cdots,K_1$}
        \For{$r=1,2,\cdots,R$}
            \State Recover near neighbours of $q$ using $\mathcal{H}_{a,r}$
            \State Store them into $\mathcal{S}$
        \EndFor
        \For{$x_{i}\in \mathcal{S}$}  
            \State $w_{i}\leftarrow f(x_{i},q)$
            \If{  {$x_{i}\in L_{r}$ for some $r\in[R]$}}
                \State $p_i \gets \min\{\frac{1}{2^r n f_{\mathsf{KDE}}}, 1\}$
            \EndIf
        \EndFor
        \State $T_{a}\leftarrow\sum_{x_{i}\in\mathcal{S}}\frac{w_i}{p_i}$
    \EndFor
    \State \Return $\mathrm{Median}_{a \in K_1} \{T_{a}\}$
\EndProcedure
\State 
\State {\bf end data structure}
\end{algorithmic}
\end{algorithm}

The running time of \textsc{Query} procedure depends on two parts: the number of recovered points in each weight level and the recovery time of LSH.

We first prove the expected number of recovered points. 

\begin{lemma}[expected number of points in level sets, informal version of Lemma~\ref{lem:expect_number_points_level_set_formal}]\label{lem:expect_number_points_level_set}
Given a query $q\in\R^d$ and fix $r\in[R]$. For any $i\in[R+1]$, weight level $L_i$ contributes at most $1$ point to the hash bucket of query $q$.
\end{lemma}

Next, we show the running time for LSH to recover points.

\begin{lemma}[running time for recovering points given a query, informal version of Lemma~\ref{lem:recover_point_from_q_formal}]\label{lem:recover_point_from_q}
Given a query $q\in\R^d$ and $L,R,k\in \mathbb{N}_+$, the \textsc{Recover} of the data-structure \textsc{LSH} runs in (expected) time
\begin{align*}
    O(Lkn^{o(1)}+LR)
\end{align*}
\end{lemma}

Combining the two lemmas above, we can prove the total running time of \textsc{Query} in \textsc{DynamicKDE} structure.

\begin{lemma}[Query part of Theorem~\ref{thm:main_result}, informal version of Lemma~\ref{lem:dynamic_KDE_query_formal}]\label{lem:dynamic_KDE_query}
Given a query $q\in\R^d$, the \textsc{Query} of the data-structure \textsc{DynamicKDE} (Algorithm~\ref{alg:dynamic_KDE_query}) runs in (expected) time 
\begin{align*}
    O(\epsilon^{-2}n^{o(1)}\log(1 / f_{\mathsf{KDE}})\cdot f_{\mathsf{KDE}}^{-o(1)}\cdot\mathrm{cost}(K)\log^3 n).
\end{align*}
\end{lemma}

%\Jiehao{I copy the LSH data structure here} \Zhao{Ok, it looks good to me.}

%% file: adversary.tex
\section{Robustness to Adversary}\label{sec:adversary}

In this section, we will turn the \textsc{Query} algorithm into a robust one. In other words, we want the following thing to happen with high probability: the algorithm responds to all query points correctly. 
We achieve this goal by taking three steps. We start with constant success probability for the \textsc{Query} procedure, which we have proved in the previous section. 

In the first step, we boost this constant probability to a high probability by applying the median technique. We note that the current algorithm succeeds with high probability only for one fixed point but we want it to respond to arbitrary query points correctly. 

It is not an easy task to generalize directly from a fixed point to infinite points in the whole space. Thus we take a middle step by introducing unit ball and $\epsilon$-net. We say a unit ball in $\R^d$ is a collection of points whose norm is less than or equal to $1$.  An $\epsilon$-net is a finite collection of points, called net points, that has the ``covering'' property. To be more specific, the union of balls that centered at net points with radius $\epsilon$ covers the unit ball. In the second step, we show that given a net of the unit ball, we have the correctness on all net points.

Finally, we show the correctness of the algorithm from net points to all points in the whole space. Then we obtain a robust algorithm.

{\bf Starting Point} In Section~\ref{sec:correctness}, we have already obtained a query algorithm with constant success probability for a fixed query point.

\begin{lemma}[Starting with constant probability]\label{lem:single_estimator}
Given $\epsilon \in (0,0.1)$, a query point $q \in \R^d$ and a set of data points $X = \{ x_{i}\}_{i=1}^{n} \subset \R^d$, let  
% \begin{align*} 
$
f_{\mathsf{KDE}}^*(q) := \frac{1}{|X|} \sum_{x \in X} f(x,q)
$
% \end{align*}
be
an estimator $\mathcal{D}$ can answer the query which satisfies:
\begin{align*}
(1-\epsilon) \cdot f_{\mathsf{KDE}}^*(q) \leq \mathcal{D}.\textsc{query}(q, \epsilon) \leq (1 + \epsilon)\cdot f_{\mathsf{KDE}}^*(q)
\end{align*}
with probability $0.9$.
\end{lemma}

% \vspace{-4mm}\Junze{What is the purpose of this?}
\paragraph{Boost the constant probability to high probability.}
 
Next, we begin to boost the success probability by repeating the query procedure and taking the median output.

\begin{lemma}[Boost the constant probability to high probability]\label{lem:fixed_points}
Let $\delta_1 \in (0,0.1)$ denote the failure probability. Let $\epsilon \in (0,0.1)$ denote the accuracy parameter.
Given $L = O( \log(1/\delta_1) )$  estimators $\{\mathcal{D}_j\}_{j=1}^{L}$. For each fixed query point $q \in \R^d$, the median of queries from $L$ estimators satisfies that:
\begin{align*}
     (1-\epsilon) \cdot f_{\mathsf{KDE}}^*(q) 
    \leq & ~ \mathrm{Median}(\{\mathcal{D}_j.\textsc{query}(q, \epsilon)\}_{j=1}^{L}) \\
    \leq & ~ (1 + \epsilon)\cdot f_{\mathsf{KDE}}^*(q)
\end{align*}
with probability $1 - \delta_1$.
\end{lemma}

\paragraph{From each fixed point to all the net points.}
 
So far, the success probability of our algorithm is still for a fixed point. We will introduce $\epsilon$-net on a unit ball and show the high success probability for all the net points. 

\begin{fact}\label{fac:number_of_net_points}
Let $N$ denote the $\epsilon_0$-net of 
%\begin{align*}
$
\{ x \in \R^d ~|~ \| x \|_2 \leq 1 \}.
$
%\end{align*}
We use $|N|$ to denote the number of points in $N$. Then $|N|\leq (10/\epsilon_0)^d$.
\end{fact}

This fact shows that we can bound the size of an $\epsilon$-net with an inverse of $\epsilon$. We use this fact to conclude the number of repetitions we need to obtain the correctness of \textsc{Query} on all net points.

\begin{lemma}[From each fixed points to all the net points]\label{lem:net_points}
Let $N$ denote the $\epsilon_0$-net of 
%\begin{align*} 
$
\{ x \in \R^d ~|~ \| x \|_2 \leq 1 \}.
$
%\end{align*}
We use $|N|$ to denote the number of points in $N$. Given $L = \log(|N|/\delta)$  estimators $\{\mathcal{D}_j\}_{j=1}^{L}$. 
With probability $1 - \delta$, we have: for all $q \in N$, the median of queries from $L$ estimators satisfies that:
\begin{align*}
   (1-\epsilon) \cdot f_{\mathsf{KDE}}^*(q) 
   \leq & ~ \mathrm{Median}(\{\mathcal{D}_j.\textsc{query}(q, \epsilon)\}_{j=1}^{L}) \\
   \leq & ~ (1 + \epsilon)\cdot f_{\mathsf{KDE}}^*(q).
\end{align*}
\end{lemma}

\paragraph{From net points to all points.}
With Lemma~\ref{lem:net_points}, we are ready to extend the correctness for net points to the whole unit ball. We demonstrate that all query points  
$\| q \|_2 \leq 1$ can be answered approximately with high probability in the following lemma.

\begin{lemma}[From net points to all points]\label{lem:from_net_points_to_all_points}
Let $\epsilon \in (0,0.1)$. Let ${\cal L} \geq 1$. Let $\delta \in (0,0.1)$. Let $\tau \in [0,1]$. 
Given $L = O(\log(( \mathcal{L}/\epsilon \tau )^d/\delta))$  estimators $\{\mathcal{D}_j\}_{j=1}^{L}$, with probability $1 - \delta$, for all query points $\|p\|_2 \leq 1$,  we have the median of queries from $L$ estimators satisfies that: $\forall \| p\|_2 \leq 1$
\begin{align*}
   % : \\
    (1-\epsilon) \cdot f_{\mathsf{KDE}}^*(p) \leq & ~ \mathrm{Median}(\{\mathcal{D}_j.\textsc{query}(q, \epsilon)\}_{j=1}^{L}) \\
    \leq & ~ (1 + \epsilon)\cdot  f_{\mathsf{KDE}}^*(p).
\end{align*}
where $q$ is the closest net point of $p$.

\end{lemma}

Thus, we obtain an algorithm that could respond to adversary queries robustly.

%% file: conclusion.tex
%\section{Limitation}
%\Zhaozhuo{Limitation below.}

\section{Conclusion}\label{sec:conclusion}
 
Kernel density estimation is an important problem in machine learning. It has wide applications in similarity search and nearest neighbor clustering. Meanwhile, in many modern scenarios, input data can change over time, and queries can be provided by adversaries. In these scenarios, we need to build adaptive data structures such that incremental changes in the input data do not require our data structure to go through costly re-initialization. Also, queries provided by adversaries do not reduce the accuracy of the estimation.  We call this problem the adaptive kernel density estimation.

We present the first such adaptive data structure design for kernel density estimation. Our data structure is efficient. It only requires subquadratic spaces. Each update to the input data only requires sublinear time, and each query can finish in sublinear time.   
It should be observed that the trade-off between efficiency and effectiveness persists in our proposed algorithms. Ordinarily, to augment the execution speed, a slight compromise on the algorithm's performance becomes inevitable. Yet, we assert that the introduction of our groundbreaking data structures pushes the boundaries of this trade-off. 
%We remark that implementing our algorithm would have carbon release to the environment. However, we wish to reduce the repetitive experiments with our analysis in theory. 
% This work is purely theoretical result, we are not aware of any negative societal impact.
%\Zhaozhuo{Negative Impact: }

\section{Impact Statement} 

Our algorithm's implementation, it must be noted, could potentially contribute to environmental carbon emissions. That said, we aim to decrease the repetition of experiments by focusing our efforts on theoretical analysis.

%% file: app_preli.tex
\paragraph{Roadmap.}

Section~\ref{app:preli} presents the basic definitions and lemmas. 
Section~\ref{app:technical} presents the technical claims for the proof of our main result. 
Section~\ref{app:data} consists of four subsections: Initialize part, Update part, Query part, and LSH part of our data structure. Each section presents the corresponding algorithm and proof of running time. 
We provide a proof sketch for correctness in Section~\ref{sec:correctness}.  
Section~\ref{app:correctness} presents the detailed proof of the correctness of our data structure. 
%We provide a proof sketch for how to make our algorithm robust to adversary queries in Section~\ref{sec:adversary}. 
Section~\ref{app:adversary} presents how to change our algorithm into an adaptive algorithm. Section~\ref{app:lipschitz} presents the Lipschitz property of the kernel function.

\section{Preliminary}\label{app:preli}

The goal of this subsection is to introduce some basic Definitions (Section~\ref{app:preli:definition}) and Lemmas (Section~\ref{app:preli:lemma}) that will be used to prove the main result.

\subsection{Definitions}\label{app:preli:definition}

We start by recalling the definition of geometric weight level.
\begin{definition}[Restatement of Definition~\ref{def:level} Geometric Weight Levels]\label{def:level_re}
Fix $R \in\mathbb{N}_+$ and $q\in\R^d$. 
We define
\begin{align*}
    w_i:=f(x_i,q)
\end{align*}

For any fix $r \in[R]:=\{1,2,\cdots,R\}$, we define  
\begin{align*}
    L_{r}:=\{x_{i}\in X ~|~ w_{i}\in(2^{-r+1},2^{-r}]\}
\end{align*}

We define the corresponding distance levels as
\begin{align*}
    z_r:=\max_{\mathrm{s.t.}f(z)\in(2^{-r},2^{-r+1}]}z.
\end{align*}
where $f(z):=f(x,q)$ for $z=\|x-q\|_2$.  

In addition, we define $L_{R+1}:=P\setminus \bigcup_{ r \in [R] } L_{r}$
\end{definition}

We restate the definition and some properties of Locality Sensitive Hashing.

\begin{definition}[Restatement of Definition~\ref{def:LSH_family}, Locally Sensitive Hash]\label{def:LSH_family_re}
A family $\mathcal{H}$ is called $(p_\mathrm{near},p_\mathrm{far},z,c)$-sensitive  where $p_\mathrm{near},p_\mathrm{far}\in [0,1],z \in\mathbb{R},c\geq 1$, if for any $x,q\in\mathbb{R}^d$:
\begin{itemize}
    \item $\Pr_{h\sim\mathcal{H}}[h(x)=h(q)~|~\| x-q\|_2\leq r]\geq p_\mathrm{near}$
    \item $\Pr_{h\sim\mathcal{H}}[h(x)=h(q)~|~\| x-q\|_2\geq cr]\leq p_{\mathrm{far}}$
\end{itemize}

\end{definition}

\subsection{Lemmas}\label{app:preli:lemma}

\begin{lemma}[Lemma 3.2 in page 6  
of \cite{ai06}]\label{lem:p_app}
Let $(a, b) \in \R^d \times \R^d $. Fixed $z > 0$, there is  
a hash family $\mathcal{H}$ such that, if $p_\mathrm{near}:=p_1(z):=\Pr_{h\sim\mathcal{H}}[h(a)=h(b)~|~\| a-b\|_2\leq z]$ and $p_\mathrm{far}:=p_2(z,c):=\Pr_{h\sim\mathcal{H}}[h(a)=h(b)~|~\| a-b\|_2\geq cz]$, 
% \Junze{The above 2 places: $\| a-b\|$ should be $\| a-b\|_2$}\Junze{Done.}
then
\begin{align*}
    \rho :=\frac{\log {1}/{p_\mathrm{near}}}{\log {1}/{p_\mathrm{far}}}\leq \frac{1}{c^2}+O(\frac{\log t}{t^\frac{1}{2}})
\end{align*}
for any $c\geq 1, t > 0$, where $p_\mathrm{near}\geq e^{-O(\sqrt{t})}$ and it requires $dt^{O(t)}$ time to evaluate.

\end{lemma}

\begin{remark}\label{rmk:p_near_app}
We find an upper bound for our definition of $\rho$ and evaluation time for hashing. For the rest part, we denote $t=\log^{\frac{2}{3}}n$. Thus we obtain $n^{o(1)}$ evaluation time and $\rho = \frac{1}{c^2}+o(1)$. Since $c=O(\log^\frac{1}{7}n)$, we have
\begin{align*}
    \frac{1}{\frac{1}{c^2}+O(\frac{\log t}{t^\frac{1}{2}})}=c^2(1-o(1)). \footnote{The above three $o(1)$ can be $\frac{\log \log^\frac{2}{3}n}{\log ^\frac{1}{3}n}, \frac{\log \log^\frac{2}{3}n}{\log ^\frac{1}{3}n}, \frac{\log \log^\frac{2}{3}n}{\log ^\frac{1}{21}n}$ respectively.}
\end{align*}
\end{remark}

In the next lemma, we show the existence of an LSH family that can separate the near points and the far points from the query with high probability.

\begin{lemma}[probability bound for separating points in different level sets, formal version of Lemma~\ref{lem:LSH}]\label{lem:LSH_formal}

Given kernel function $f$, we have corresponding weight level sets $L_r$'s and distance levels $z_r$'s (Definition~\ref{def:level}). Given query $q\in\R^d$ and integer $i\in [R+1]$, $r\in[R]$ satisfying $i>r$, let $x\in L_r$, $x^\prime \in L_i, c_{i,r}:=\min\{\frac{z_{i-1}}{z_r},\log^{1/7}n\}$. We set up an Andoni-Indyk LSH family $\mathcal{H}$ (Definition~\ref{def:LSH_family}) with near distance $z_r$. 

We define
\begin{align*}
    p_{\mathrm{near},r}&~:=\Pr_{h\sim\mathcal{H}}[h(x)=h(q)~|~\| x-q\|_2\leq z]\\
    p_{\mathrm{far},r}&~:=\Pr_{h\sim\mathcal{H}}[h(x)=h(q)~|~\| x-q\|_2\geq cz]
\end{align*}

Then the following inequalities holds for any integer $k \in \mathbb{N}_+$
\begin{enumerate}
    \item $\Pr_{h^*\sim \mathcal{H}^k}[~h^*(x)=h^*(q)~]\geq p_{\mathrm{near},r}^k$
    \item $\Pr_{h^*\sim \mathcal{H}^k}[~h^*(x^\prime)=h^*(q)~]\leq p_{\mathrm{near},r}^{kc_{i,r}^2(1-o(1))}$
\end{enumerate}
 
\end{lemma}
\begin{proof}
Since $x\in L_r$, by Lemma~\ref{def:level}, we have
\begin{align}\label{eq:p_r_j}
    \|x-q\|_2\leq z_r
\end{align}
For $x^\prime \in L_i$, since we assume the $f$ is decaying radial kernel, we have
\begin{align}\label{eq:p_prime_r_j}
    \|x^\prime-q\|_2\geq z_{i-1}\geq c_{i,r}z_r
\end{align}

where the first step follows from Definition~\ref{def:level}, the last step follows from $c_{i,r} \geq \tilde{c}_r$. 

By Lemma~\ref{lem:p} and Eq.~\eqref{eq:p_r_j}, Eq.~\eqref{eq:p_prime_r_j}, we have
\begin{enumerate}
    \item $\Pr_{h\sim \mathcal{H}}[h(x)=h(q)]\geq p_{\mathrm{near},r}$
    \item $\Pr_{h\sim \mathcal{H}}[h(x^\prime)=h(q)]\leq p_{\mathrm{far},r}$
\end{enumerate}

By remark~\ref{rmk:p_near}, we have
\begin{align}\label{eq:p_far_leq_p_near}
    p_{\mathrm{far},r}\leq p_{\mathrm{near},r}^{c_{i,r}(1-o(1))}
\end{align}

Then for any integer $k>1$, we have
\begin{align*}
    \Pr_{h^*\sim \mathcal{H}^k}[h^*(x)=h^*(q)]&~\geq~p_{\mathrm{near},r}^k\\
\Pr_{h^*\sim \mathcal{H}^k}[h^*(x^\prime)=h^*(q)]&~\leq~p_{\mathrm{far},r}^{k}
\end{align*}

By Eq.~\eqref{eq:p_far_leq_p_near}, we obtain the final result
\begin{enumerate}
    \item $\Pr_{h^*\sim \mathcal{H}^k}[h^*(x)=h^*(q)]\geq p_{\mathrm{near},r}^k$
    \item $\Pr_{h^*\sim \mathcal{H}^k}[h^*(x^\prime)=h^*(q)]\leq p_{\mathrm{near},r}^{kc_{i,r}^2(1-o(1))}$
\end{enumerate}

Thus, we complete the proof.
\end{proof}

%% file: app_technical.tex
\section{Technical Claims}\label{app:technical}

 In Section~\ref{app:technical:lem_upper_bound_geometric}, we show how to bound the size of geometric weight levels. In Section~\ref{app:technical:lem_lower_bound_recovered_point}, we explain how show the probability for sampled point recovery. In Section~\ref{app:technical:lem_upper_bound_recovered_point}, we give an expectation bound the number of recovered points.
   
\subsection{Size of geometric weight levels}\label{app:technical:lem_upper_bound_geometric}

The goal of this section is to prove Lemma~\ref{lem:upper_bound_geometric}.

\begin{lemma}[Sizes of geometric weight levels]\label{lem:upper_bound_geometric}
Given $r\in[R]$, we have
\begin{align*}
    |L_r|\leq 2^r n f_{\mathsf{KDE}}^*\leq 2^r n f_{\mathsf{KDE}}.
\end{align*}
\end{lemma}

\begin{proof}
For any fix $r\in [R]$, $x,q\in\mathbb{R}^d$, we have
\begin{align*}
    n \cdot f_{\mathsf{KDE}} \geq & ~ n \cdot f_{\mathsf{KDE}}^* \\
    = & ~ \sum\limits_{x \in X}f(x,q) \\
    \geq & ~ \sum\limits_{p\in L_r}f(x,q)\\ 
    \geq & ~  |L_r| 2^{-r}
\end{align*}
where the first step follows from $f_{\mathsf{KDE}} \geq f_{\mathsf{KDE}}^*$, the second step follows from Definition~\ref{def:dynamic_KDE}, the third step follows from shrinking the number of summands, the last step follows from Definition~\ref{def:level}.

Thus, we complete the proof.
\end{proof}

\subsection{Probability for sampled point recovery}\label{app:technical:lem_lower_bound_recovered_point}

The goal of this section is to prove Lemma~\ref{lem:lower_bound_recovered_point}.
 
\begin{lemma}[Probability for sampled point recovery]\label{lem:lower_bound_recovered_point}
Suppose that we invoke \textsc{DynamicKDE.Initialize}. Suppose when $a=a^*$ and $r=r^*$, we sample a point $x \in L_{r^*}$. Given a query $q$, we invoke \textsc{DynamicKDE.Query}. With probability at least $1-\frac{1}{n^{10}}$, $H_{a^*,r^*}$ recovered $x$.
 
\end{lemma}

\begin{proof}
By Lemma~\ref{lem:LSH_formal} we have
\begin{align*}
    \Pr_{h^*\sim\mathcal{H}^k}[h^*(x)=h^*(q)]\geq p_{\mathrm{near},r^*}^k
\end{align*}

Now note that in $\textsc{LSH.Recover}$ (line~\ref{lin:evaluate_recover}) procedure, we repeat this process for
\begin{align*}
    K_{2,r^*}=100\log (n) p_{\mathrm{near},r^*}^{-k}
\end{align*}
times. Thus, for any sampled point $ p \in L_{r^*}$, it is recovered in one of the repetitions of phase $r=r^*$, with probability at least $1-\frac{1}{n^{10}}$.
 
\end{proof}

\subsection{Number of recovered points in expectation}\label{app:technical:lem_upper_bound_recovered_point}

The goal of this section is to prove Lemma~\ref{lem:upper_bound_recovered_point}
\begin{lemma}[Upper bound on number of recovered points in expectation]\label{lem:upper_bound_recovered_point}
Fix a query $q\in\mathbb{R}^d$. We define $R:=\lceil \log \frac{1}{f_{\mathsf{KDE}}} \rceil$. Fix $r \in [R]$, we define $ p:=p_{\mathrm{near},r}$. For each $(i,r)\in [R] \times [R]$, we define $c_{i,r}:=\min\{\frac{z_{i-1}}{z_r},\log^\frac{1}{7}n\}$.  
There exists $k\in\mathbb{N}_+$ 
\begin{align*}
    k:=k_r:=\frac{1}{\log (1/p)}\max\limits_{l\in\{r+1,\cdots,R+1\}}\lceil\frac{l-r}{c_{l,r}^2(1-o(1))}\rceil.
\end{align*}
such that for any $i>j$
\begin{align*}
    \E_{h^*\sim \mathcal{H}^{k}}[|\{x^\prime \in L_i:h^*(x^\prime)=h^*(q)\}|]=O(1)
\end{align*}
 
\end{lemma}
\begin{proof}
By Lemma~\ref{lem:LSH_formal} we have
\begin{align*}
    \Pr_{h^*\sim\mathcal{H}^k}[h^*(x)=h^*(q)]\leq p^{k c_{i,r}^2(1-o(1))}
\end{align*}
where $c_{i,j}:=\min\{\frac{r_{i-1}}{r_j},\log^\frac{1}{7}n\},p:=p_{\mathrm{near},j}\in (0,1)$(remark~\ref{rmk:p_near}).
\begin{align*}
    & ~\E_{h^*\sim \mathcal{H}^{k}}[|x^\prime \in L_i:h^*(x^\prime)=h^*(q)|]\\ 
    \leq & ~ 2^i n f_{\mathsf{KDE}}\cdot \frac{1}{2^{r}n f_{\mathsf{KDE}}}\Pr_{h^*\sim\mathcal{H}^k}[h^*(x)=h^*(q)]\\ 
    \leq & ~ 2^{i-r}\cdot p^{k c_{i,r}^2(1-o(1))}
\end{align*}
where the first step follows from lemma~\ref{lem:upper_bound_geometric} and sampling probability (Algorithm~\ref{alg:dynamic_KDE_initialize} line~\ref{lin:sample_prob}), the second step follows from Lemma~\ref{lem:LSH}.

Note that for $i>j$, we have 
\begin{align}\label{eq:lower_bound_k}
    k\geq \frac{1}{\log \frac{1}{p}}\lceil\frac{i-r}{c_{i,r}^2(1-o(1))}\rceil
\end{align}

Then
\begin{align*}
    &~2^{i-r}\cdot p^{k c_{i,r}^2(1-o(1))}\\
    =&~2^{i-r}\cdot 2^{\log(p)\cdot k c_{i,r}^2(1-o(1))}\\
    \leq &~2^{i-r}2^{\log (p) \cdot \frac{1}{\log (\frac{1}{p})}\lceil\frac{i-r}{c_{i,r}^2(1-o(1))}\rceil c_{i,r}^2(1-o(1))}\\
    \leq &~2^{i-r+r-i} \\
    =&~1  
\end{align*}
where the first step follows from rewriting in exponential form, the second step follows from Eq.~\eqref{eq:lower_bound_k}  
, the third step follows from canceling the same term in both numerator and denominator, and the last step follows from canceling $i$ and $r$.

Thus, we complete the proof.
\end{proof}

%% file: app_data.tex
\section{Our Data Structures}\label{app:data}
In this section, we describe our data structures in detail. Starting with the initialize part in Section~\ref{app:data_structure_init}, we state the result of space storage and running time for Initialize in \textsc{DynamicKDE}. In Section~\ref{app:data_structure_update}, we demonstrate the running time for the update part in our data structure. Section~\ref{app:data_structure_query} presents the running time for the query procedure. Finally, we study the LSH data structure in Section~\ref{app:data_structure_lsh}. It is an important member in the \textsc{DynamicKDE} structure and fundamental to the implementation of all three procedures above.

\subsection{Initialize part of data structure}\label{app:data_structure_init}

In this section, we describe the space storage and running time of \textsc{Initialize} part of our data structure \textsc{DynamicKDE}.

We start by showing the space storage of \textsc{LSH} structure.

\begin{lemma}[Space storage of \textsc{LSH}, formal version of Lemma~\ref{lem:LSH_storage}]\label{lem:LSH_storage_formal}
Given data set $\{x_i\}_{i\in[n]}\subset\R^d$, parameter $L,k\in\mathbb{N}_+$, the \textsc{Initialize} (Algorithm~\ref{alg:LSH_public_app}) of the data-structure \textsc{LSH} uses space
\begin{align*}
    O(Lkdn^{o(1)}+Ln)
\end{align*}
\end{lemma}
\begin{proof}
The space storage comes from two parts: \textsc{ChooseHashFunc} and \textsc{ConstructHashTable}.

{\bf Part 1.} \textsc{ChooseHashFunc}(line~\ref{lin:choose_hash_func}) takes $L,k$ as input. 

It has a for loop with $L$ iterations.

In each iteration, it samples $k$ functions(line~\ref{lin:LSH_sample_k_functions}) from hash family $\mathcal{H}$ to create $\mathcal{H}_l$, which uses $O(kdn^{o(1)})$ space.

Thus the total space usage of \textsc{ChooseHashFunc} is $L\cdot O(kdn^{o(1)})=O(Lkdn^{o(1)})$.

{\bf Part 2.} \textsc{ConstructHashTable}(line~\ref{lin:construct_hash_table}) takes data set $\{x_i\}_{i\in[n]}$ and parameter $L$ as input.

It has two recursive for loops.
\begin{itemize}
    \item The first for loop repeats $L$ iterations.
    \item The second for loop repeats $n$ iterations.
\end{itemize}

The space storage of the inner loop comes from line~\ref{lin:insert} and line~\ref{lin:aggregate_hash_table}, which is $O(1)$.

Thus the total space storage of \textsc{ConstructHashTable} is $L\cdot n\cdot O(1)=O(Ln)$.

The final space storage of \textsc{Initialize} is 
\begin{align*}
    &~{\bf Part 1}+{\bf Part 2}\\
    =&~O(Lkdn^{o(1)}+Ln)
\end{align*}

Thus, we complete the proof.

\end{proof} 

Using the above lemma, we state the space storage of our \textsc{DynamicKDE} structure.

\begin{lemma}[Space storage part of Theorem~\ref{thm:main_result}, formal version of Lemma~\ref{lem:init}]\label{lem:init_formal}
The \textsc{Initialize} of the data structure \textsc{DynamicKDE} (Algorithm~\ref{alg:dynamic_KDE_initialize}) uses space
\begin{align*}
    O(& \epsilon^{-2}(\frac{1}{f_{\mathsf{KDE}}})^{o(1)}\cdot\log(1/f_{\mathsf{KDE}})\cdot \mathrm{cost}(K)
    \\ & \cdot\log^2 n\cdot(\frac{1}{f_{\mathsf{KDE}}}+n^{o(1)}\cdot\log^2 n))
\end{align*}
\end{lemma}

\begin{proof}
The space storage mainly comes from $K_1\cdot J$ copies of $\mathcal{H}$. 

Now let's consider the space storage of $\mathcal{H}$. By Lemma~\ref{lem:LSH_storage}, we replace $\{x_i\}_{i\in[n]},L,k$ with $P_j$ (line~\ref{lin:sample}), $K_{2,j}$ (line~\ref{lin:K_2}), $k_j$ (line~\ref{lin:k_j}) respectively. We have $|P_j|=O(\frac{1}{f_{\mathsf{KDE}}}),K_{2,j}=O(\mathrm{cost}(K)\cdot\log n)$ and $k_j=O(\log n)$. Thus the total space usage of $\mathcal{H}$ is
\begin{align}\label{eq:LSH_space_storage}
    & ~ O(Lkdn^{o(1)}+Ln)
    \\ 
    = & ~O(\mathrm{cost}(f)n^{o(1)}\cdot\log^3 n+ \mathrm{cost}(f)(\frac{1}{f_{\mathsf{KDE}}})\cdot \log n)
\end{align}

The total space storage of \textsc{Initialize} of the data structure \textsc{DynamicKDE} is
\begin{align*}
    &~K_1\cdot J\cdot O(Lk+Ln)\\
    =&~O(K_1\cdot J\cdot \mathrm{cost}(K)\log n\cdot(\frac{1}{f_{\mathsf{KDE}}}+\log n))\\
    =&~O(\epsilon^{-2}(\frac{1}{f_{\mathsf{KDE}}})^{o(1)}\cdot\log(1/f_{\mathsf{KDE}})\cdot \mathrm{cost}(K)
    \\ & \cdot\log^2 n\cdot(\frac{1}{f_{\mathsf{KDE}}}+n^{o(1)}\cdot\log^2 n))
\end{align*}
where the first step follows from Eq.~\eqref{eq:LSH_space_storage}, the last step follows from $K_1=O(\epsilon^{-2} (\frac{1}{f_{\mathsf{KDE}}})^{o(1)}\cdot\log n)$ and $J=O(\log(1/f_{\mathsf{KDE}}))$.

Thus, we complete the proof.
\end{proof}

Next, we show an upper bound on running time for \textsc{Initilize} in \textsc{LSH} data structure. 

\begin{lemma}[Upper bound on running time of \textsc{Initialize} of the data-structure \textsc{LSH}, formal version of Lemma~\ref{lem:LSH_initialize_upper_bound} ]\label{lem:LSH_initialize_upper_bound_formal}

Given input data points $\{x_i\}_{i\in[n]}\subset \mathbb{R}^d$, parameters $k,L\in \mathbb{N}_+$, LSH parameters $p_{\mathrm{near}},p_{\mathrm{far}} \in [0,1],c\in[1,\infty), r\in \R_+$ and kernel $f$, the \textsc{Initialize} of the data-structure \textsc{LSH}(Algorithm~\ref{alg:LSH_public_app}) runs in time
\begin{align*}
    O(L\cdot(kdn^{o(1)}+dn^{1+o(1)}+n\log n))
\end{align*}
\end{lemma}
\begin{proof}
This procedure consists of three parts

{\bf Part 1.} We invoke \textsc{ChooseHashTable} procedure with parameters $k,L$ (line~\ref{lin:LSH_intialize_choose_hash_func}). The \textsc{ChooseHashTable} procedure has one for loop with L iterations.

Now let's consider the running time in line~\ref{lin:LSH_sample_k_functions}, which is the running time in each iteration. In line~\ref{lin:LSH_sample_k_functions}, we sample $k$ hash functons from hash family $\mathcal{H}$, which takes $O(k\cdot dn^{o(1)})$ time.  

Thus the total running time for {\bf Part 1} is
\begin{align*}
    O(Lkdn^{o(1)})
\end{align*}

{\bf Part 2.} We invoke \textsc{ConstructHashTable} procedure with data set $\{x_i\}_{i\in[n]}$. This procedure has two recursive for loops.
\begin{itemize}
    \item The first loops repeat $L$ iterations
    \item The second loop repeats $n$ iterations
\end{itemize}

Now let's consider the running time from line~\ref{lin:find_bucket_insert_element} to line~\ref{lin:aggregate_hash_table}, which is the time for each inner loop. 
\begin{itemize}
    \item line~\ref{lin:find_bucket_insert_element}: We first evaluate $\mathcal{H}_l(x_i)$, which takes $O(dn^{o(1)})$. Then we insert $x_i$ in the bucket $\mathcal{H}_l(x_i)$, which takes $O(\log n)$ time.
    \item line~\ref{lin:aggregate_hash_table} takes $O(1)$ time.
\end{itemize}

The running time from line~\ref{lin:find_bucket_insert_element} to line~\ref{lin:aggregate_hash_table} is $O(dn^{o(1)}+\log n)$

The total running time for {\bf Part 2} is
\begin{align*}
    O(Ln\cdot(dn^{o(1)}+\log n))
\end{align*}

{\bf Putting it all together.} We prove that the \textsc{Initialize} of the data-structure \textsc{LSH}(Algorithm~\ref{alg:LSH_public_app}) runs in time
\begin{align*}
    &~{\bf Part 1}+{\bf Part 2}\\
    =&~O(Lkdn^{o(1)})+O(Ln\cdot(dn^{o(1)}+\log n))\\
    =&~O(L\cdot(kdn^{o(1)}+dn^{1+o(1)}+n\log n))
\end{align*}

Thus, we complete the proof.

\end{proof}

Combining the results above, we can demonstrate the running time of \textsc{Initialize} in \textsc{DynamicKDE} in the following lemma.

\begin{lemma}[The initialize part of Theorem~\ref{thm:main_result}, formal version of Lemma~\ref{lem:dynamic_KDE_initialize}]\label{lem:dynamic_KDE_initialize_formal}
Given $(f:\mathbb{R}^d\times \mathbb{R}^d \rightarrow [0,1], P\subset\mathbb{R}^d, \epsilon \in (0,1),f_{\mathsf{KDE}}\in[0,1])$, the \textsc{Initialize} of the data-structure \textsc{DynamicKDE} (Algorithm~\ref{alg:dynamic_KDE_initialize}) runs in time
\begin{align*}
    O(\epsilon^{-2}n^{1+o(1)}\cost(f)\cdot (\frac{1}{f_{\mathsf{KDE}}})^{o(1)}\log(1/f_{\mathsf{KDE}}) \cdot \log^2 n)
\end{align*}
\end{lemma}

\begin{proof}
The \textsc{Initialize} procedure has two recursive for loops.
\begin{itemize}
    \item The first loops repeat $K_1=O(\epsilon^{-2}\log (n)\cdot f_{\mathsf{KDE}}^{-o(1)})$ iterations
    \item The second loops repeats $J=O(\log \frac{1}{f_{\mathsf{KDE}}})$ iterations
\end{itemize}

Now let's consider the running time from line~\ref{lin:c_ij} to line~\ref{lin:LSH_initialize_in_KDE}, which is the running time of the inner loop.
\begin{itemize}
    \item line~\ref{lin:c_ij} to line~\ref{lin:sample_prob} takes $O(\log(1/f_{\mathsf{KDE}}))$ time.
    \item line~\ref{lin:sample} takes $O(n)$ time.
    \item line~\ref{lin:LSH_initialize_in_KDE}:
    By Lemma~\ref{lem:LSH_initialize_upper_bound}, we replace $L$ with $K_{2,j}=O(\mathrm{cost}(K)\cdot \log n)$ and $k$ with $k_j=O(\log n)$ .
    
    Thus the running time of this line is
    \begin{align*}
        &~O(L\cdot(kdn^{o(1)}+dn^{1+o(1)}+n\log n)\\
        =&~O(\mathrm{cost}(K)\cdot \log n\cdot(n^{o(1)}\cdot\log^2  n+n^{1+o(1)}\cdot\log  n+n\cdot\log n))\\
        =&~O(n^{1+o(1)}\mathrm{cost}(K)\cdot\log^2 n)
    \end{align*}
    
\end{itemize}
where the first step follows from $K_{2,j}=O(\mathrm{cost}(K)\cdot \log n)$, $d=O(\log n)$ and $k_j=O(\log n)$, the second step follows from $O(n\log n)=O(n^{1+o(1)})$.

The running time from from line~\ref{lin:c_ij} to line~\ref{lin:LSH_initialize_in_KDE} is
\begin{align*}
    &~O(\log(1/f_{\mathsf{KDE}}))+O(n)+O(n^{1+o(1)}\mathrm{cost}(K)\log n)\\
    =&~O(n^{1+o(1)}\mathrm{cost}(K)\log^2 n)
\end{align*}

The final running time for \textsc{Initialize} procedure is
\begin{align*}
    &~K_1\cdot J \cdot O(n^{1+o(1)}\mathrm{cost}(K)\cdot\log^2 n)\\
    =&~O(\epsilon^{-2}n^{1+o(1)}\cost(f)\cdot (\frac{1}{f_{\mathsf{KDE}}})^{o(1)}\cdot\log(1/f_{\mathsf{KDE}}) \cdot \log^3 n)
\end{align*}
where we use $K_1=O(\epsilon^{-2}\cdot f_{\mathsf{KDE}}^{-o(1)}\cdot\log n)$ and $J=O(\log(1/f_{\mathsf{KDE}}))$. 

Thus, we complete the proof.
\end{proof}

\begin{algorithm}[!ht]\caption{Dynamic KDE, members and initialize part}\label{alg:dynamic_KDE_initialize}
\begin{algorithmic}[1]
\State {\bf data structure} \textsc{DynamicKDE}      \Comment{Theorem~\ref{thm:main_result}}
\State {\bf members}
    \State \hspace{4mm} $d \in \mathbb{N}_+$ \Comment{Dimension of data point}
    \State \hspace{4mm} For ${i\in[n]}$,$x_i \in \mathbb{R}^d$ \Comment{dataset $X$}
    \State \hspace{4mm} $K_1 \in \mathbb{N}_+$ \Comment{Number of repetitions}
    \State \hspace{4mm} $R \in \mathbb{N}_+$
    \State \hspace{4mm} For $a \in [K_1]$, $\wt{P}_a \subset \mathbb{R}^d$ \Comment{Sampled data points}
    \State \hspace{4mm} $K_2 \in \mathbb{N}_+$
    \State \hspace{4mm} For $a \in [K_1]$, $r \in [R]$, $H_{a,r}\in \textsc{LSH}$ \label{lin:instance_LSH_app} \Comment{Instances from \textsc{LSH} class}
\State {\bf end members}

\State

\Procedure{\textsc{Initialize}}{$X \subset \R^d,\epsilon \in (0,1),f_{\mathsf{KDE}} \in [0,1]$} \Comment{Lemma~\ref{lem:init}}
    \State \Comment{$f_{\mathsf{KDE}}$ is a known quantity satisfy $f_{\mathsf{KDE}}\geq f_{\mathsf{KDE}}^{*}$}
    \State \Comment{$\epsilon$ represents the precision of estimation}
    \State $K_1\leftarrow C\cdot \epsilon^{-2} \log{n}\cdot f_{\mathsf{KDE}}^{-o(1)}$\label{lin:K1}
    \State $R \leftarrow \left\lceil\log{{1}/{f_{\mathsf{KDE}}}}\right\rceil$\label{lin:J}
    \For{$a=1,2,\cdots,K_1$}
        \For{$r=1,2,\cdots,R$}
            \For{$i=r+1,\cdots,R+1$}
                \State $c_{i,r}\leftarrow\min\{\frac{z_{i-1}}{z_{r}},\log^\frac{1}{7}n\}$\label{lin:c_ij} \Comment{$z_r$ is defined in Definition~\ref{def:level}}
            \EndFor
            \State $k_r\leftarrow \max_{i \in \{r+1, \cdots, R+1\}}\frac{1}{\log \frac{1}{p}}\lceil\frac{i-r}{\tilde{c}_{i,r}(1-o(1))}\rceil$\label{lin:k_j}
            \State $p_{\mathrm{near},r}\leftarrow p(z_r)$
            \State $K_{2,r} \leftarrow 100\log{n}\cdot p_{\mathrm{near},r}^{-k_r} $\label{lin:K_2}  
            \Comment{$p_{\mathrm{near},r},p_{\mathrm{far},r}$ are defined in Lemma~\ref{lem:p}}
            \State $p_{ \mathrm{sampling} }\leftarrow \min\{\frac{1}{2^{r}nf_{\mathsf{KDE}}},1\}$ \label{lin:sample_prob}
            \State $P_r\leftarrow$ sample each element in $X$ with probability $p_{\mathrm{sampling}}$.\label{lin:sample}
            \State $\mathcal{H}_{a,r}.\textsc{Initialize}(P_r,k_r,K_{2,r})$ \label{lin:LSH_initialize_in_KDE}
        \EndFor
        \State $\tilde{P}_{a}\leftarrow$ sample each element in $X$ with probability $\frac{1}{n}$ \Comment{Store $\tilde{P}_a$}
    \EndFor
\EndProcedure
\State {\bf end data structure}
\end{algorithmic}
\end{algorithm}

\begin{algorithm}[!ht]\caption{Dynamic KDE, update part}\label{alg:dynamic_KDE_update}
\begin{algorithmic}[1]
\State {\bf data structure} \textsc{DynamicKDE} \Comment{Theorem~\ref{thm:main_result}}
\State

\Procedure{\textsc{Update}}{$v \in \R^d, f_{\mathsf{KDE}} \in [0,1], i \in [n]$} \Comment{Lemma~\ref{lem:dynamic_KDE_update}}
    \For{$a=1,2,\cdots,K_1$}
        \For{$r=1,2,\cdots,R$}
            \State $\mathcal{H}_{a,r}.\textsc{UpdateHashTable}(v,i)$\label{update_hashtable_in_dynamic_KDE}
        \EndFor
    \EndFor
    \State $x_i\leftarrow v$
\EndProcedure
\State {\bf end data structure}
\end{algorithmic}
\end{algorithm}

\subsection{Update part of data structure}\label{app:data_structure_update}
The goal of this section is to prove Lemma~\ref{lem:dynamic_KDE_update}. Our Lemma~\ref{lem:dynamic_KDE_update_formal} in this section is the formal version of Lemma~\ref{lem:dynamic_KDE_update}. We present an auxiliary Lemma~\ref{lem:hash_update_formal} and then show how to this this auxiliary lemma to prove Lemma~\ref{lem:dynamic_KDE_update_formal}.

\begin{lemma}[Update time of LSH, formal version of Lemma~\ref{lem:hash_update}]\label{lem:hash_update_formal}
Given a data point $z\in\mathrm{R}^d$ and index $i\in[n]$, the \textsc{UpdateHashTable} of the data-structure \textsc{LSH} runs in (expected) time
\begin{align*}
    O(n^{o(1)}\log(n)\cdot \cost(f)).
\end{align*}

\end{lemma}

\begin{proof}

This procedure has one for loop which repeats $L=O(\log n)$ iterations.
Now let us consider the running time from line~\ref{lin:insert} to line~\ref{lin:delete}, which is the time for each iteration.

\begin{itemize}
    \item line~\ref{lin:insert} takes $O(dn^{o(1)}\cost(f))$
    \item line~\ref{lin:delete} takes the same time as line~\ref{lin:insert}
\end{itemize}

The final running time
\begin{align*}
    &~L\cdot O(dn^{o(1)}\cost(f))\\
    =&~O(n^{o(1)}\cost(f)\cdot \log^2 n).
\end{align*}
where we use $L=O(\log n)$ and $d=O(\log n)$.

Thus, we complete the proof.

\end{proof}

\begin{lemma}[The update part of Theorem~\ref{thm:main_result}, formal version of Lemma~\ref{lem:dynamic_KDE_update}]\label{lem:dynamic_KDE_update_formal}
Given an update $z\in\R^d$ and index $i\in[n]$, the \textsc{Update} of the data-structure \textsc{DynamicKDE} (Algorithm~\ref{alg:dynamic_KDE_update}) runs in (expected) time
\begin{align*}
    O(\epsilon^{-2}n^{o(1)}\cost(f)\cdot (\frac{1}{f_{\mathsf{KDE}}})^{o(1)}\log(1/f_{\mathsf{KDE}}) \cdot \log^3 n).
\end{align*}
\end{lemma}

\begin{proof}

This algorithm has two recursive for loops
\begin{itemize}
    \item The first loops repeat $K_1=O(\epsilon^{-2}\log (n)\cdot f_{\mathsf{KDE}}^{-o(1)})$ iterations
    \item The second loops repeats $J=O(\log \frac{1}{f_{\mathsf{KDE}}})$ iterations
\end{itemize}

Now let's consider the running time in line~\ref{update_hashtable_in_dynamic_KDE}, which is the time for each inner loop.

By Lemma~\ref{lem:hash_update}, line~\ref{update_hashtable_in_dynamic_KDE} takes $O(n^{o(1)}\log(n)\cdot \cost(f))$ time. 

The final running time
\begin{align*}
    &~K_1\cdot J\cdot O(n^{o(1)}\cost(f)\cdot \log^2 n)\\
    =&~O(\epsilon^{-2}n^{o(1)}\cost(f)\cdot (\frac{1}{f_{\mathsf{KDE}}})^{o(1)}\log(1/f_{\mathsf{KDE}}) \cdot \log^3 n)
\end{align*}
where we use $K_1=O(\epsilon^{-2}\log (n)\cdot f_{\mathsf{KDE}}^{-o(1)})$ and $J=O(\log \frac{1}{f_{\mathsf{KDE}}})$.

Thus, we complete the proof.
\end{proof}

\subsection{Query part of data structure}\label{app:data_structure_query}
The goal of this section is to prove Lemma~\ref{lem:dynamic_KDE_query_formal}.
Our Algorithm~\ref{alg:dynamic_KDE_query} is for querying the approximated kernel density at point $q$, and Lemma~\ref{lem:dynamic_KDE_query_formal} specifies the running time for the query operation.
In order to prove this lemma, we list and prove a few auxiliary lemmas.

\begin{algorithm}[!ht]\caption{Dynamic KDE, query part}\label{alg:dynamic_KDE_query}
\begin{algorithmic}[1]
\State {\bf data structure} \textsc{DynamicKDE} \Comment{Theorem~\ref{thm:main_result}}
\State
 
\Procedure{\textsc{Query}}{$q\in \mathbb{R}^d, \epsilon \in (0,1),f_{\mathsf{KDE}} \in [0,1]$} 
 
    \For{$a=1,2,\cdots,K_1$}\label{lin:first_loop}
        \For{$r=1,2,\cdots,R$}\label{lin:second_loop}
            \State $\mathcal{H}_{a,r}.\textsc{Recover}(q)$\label{lin:evaluate_recover}
            \State $\mathcal{S} \leftarrow \mathcal{S} \cup (\mathcal{H}_{a,r}.\mathcal{R}\cap L_r) $\label{lin:choose_L_j_point}
        \EndFor\label{lin:end_second_loop}
        \State $\mathcal{R}_{R+1}\leftarrow$ recover points in $L_{R+1}\cap\tilde{P}_{a}$\label{lin:recover_point_J+1} \Comment{Recover by calculating $w$ directly.}
        \State $\mathcal{S} \leftarrow \mathcal{S}\cup\mathcal{R}_{R+1}$\label{lin:add_point_to_S}
        \For{$x_{i}\in \mathcal{S}$} \label{lin:third_loop} 
            \State $w_{i}\leftarrow f(x_{i},q)$
            \If{$x_{i}\in L_{r}$ for some $r\in[R]$}
                \State $p_{i}\leftarrow\min\{ \frac{1}{2^{r} n f_{\mathsf{KDE}} },1\}$
            \ElsIf{$x_{i} \in X \setminus \bigcup_{ r \in [R] } L_{r}$}
                \State $p_{i}\leftarrow\frac{1}{n}$
            \EndIf
        \EndFor\label{lin:end_third_loop}
        \State $T_{a}\leftarrow\sum_{x_{i}\in\mathcal{S}}\frac{w_i}{p_i}$ \label{lin:output_Z_a}
    \EndFor
    \State \Return $\mathrm{Median}\{T_{a}\}$
\EndProcedure
\State {\bf end data structure}
\end{algorithmic}
\end{algorithm}

We start by showing a lemma that states the expected number of points in each level set.

\begin{lemma}[expected number of points in level sets, formal version of Lemma~\ref{lem:expect_number_points_level_set}]\label{lem:expect_number_points_level_set_formal}
Given a query $q\in\R^d$ and fix $r\in[R]$. For any $i\in[R+1]$, weight level $L_i$ contributes at most $1$ point to the hash bucket of query $q$.
\end{lemma}

\begin{proof}
We consider 2 cases:

{\bf Case 1.} $i\leq r$: By lemma~\ref{lem:upper_bound_geometric}, we have $|L_i|\leq 2^inf_{\mathsf{KDE}}$. In the $r$'th phase, we sample each point in the whole data set with probability $\min\{\frac{1}{2^r nf_{\mathsf{KDE}}},1\} $ to obtain a subset $X_r$ (Algorithm~\ref{alg:dynamic_KDE_initialize} line~\ref{lin:sample}). Then
\begin{align*}
    &~\E[|\{x:x\in L_i\cap X_r\}|]\\
    \leq&~|L_i|\cdot \frac{1}{2^r n f_{\mathsf{KDE}}}\\
    \leq&~2^i n f_{\mathsf{KDE}}\cdot \frac{1}{2^r n f_{\mathsf{KDE}}}\\
    =&~2^{i-r}\\\
    \leq&~1
\end{align*}
where the first step follows from sampling probability $\min\{\frac{1}{2^r n f_{\mathsf{KDE}}},1\}$, the second step follows from $|L_i|\leq 2^i n f_{\mathsf{KDE}}$, the third step follows from canceling $n f_{\mathsf{KDE}}$, the last step follows from $i\leq r$.

Thus, there is at most $1$  
sampled point from $L_i$ in expectation.
 
Then $L_i$ contributes at most $1$ point in the bucket of query $q$ in expectation. 
 
{\bf Case 2.} $i=r+1,\cdots,R + 1$: By Lemma~\ref{lem:LSH}, we have $|L_i|\leq2^i n f_{\mathsf{KDE}}$. The sampling rate in $r$'th phase is $\min\{\frac{1}{2^r n f_{\mathsf{KDE}}},1\}$(Algorithm~\ref{alg:dynamic_KDE_initialize} line~\ref{lin:sample}). Then there are at most $2^{i-r}$ sampled points from $L_i$ in expectation. We set up LSH function such that the near distance is $z_r$ (Definition~\ref{def:level}). Also, we use (Algorithm~\ref{alg:dynamic_KDE_initialize} line~\ref{lin:k_j})  
\begin{align*}
    k:=k_r:=\frac{1}{\log \frac{1}{p}}\max\limits_{i=r+1,\cdots,R+1}\lceil\frac{i-r}{c_{i,r}(1-o(1))}\rceil.
\end{align*}
as the number of concatenations. By Lemma~\ref{lem:upper_bound_recovered_point}, $L_i$ contributes at most $1$ point in the bucket of query $q$ in expectation.

The total number of points that $L_i$ contributes to hash bucket of $q$ is $\max\{${\bf Case 1},{\bf Case 2}$\}=1$ in expectation.

Thus, we complete the proof.
\end{proof}

Next, we present the running time of \textsc{Recover} procedure in the LSH data structure, which is an important part of \textsc{Query} procedure in \textsc{DynamicKDE}.

\begin{lemma}[running time for recover points given a query, formal version of Lemma~\ref{lem:recover_point_from_q}]\label{lem:recover_point_from_q_formal}
Given a query $q\in\R^d$ and $L,R,k\in \mathbb{N}_+$, the \textsc{Recover} of the data-structure \textsc{LSH} runs in (expected) time
\begin{align*}
    O(L k n^{o(1)}+LR)
\end{align*}
\end{lemma}
\begin{proof}
The procedure has one for loop with $L$ iterations.
In each iteration, the running time consists of two parts
\begin{itemize}
    \item The evaluation of $\mathcal{H}_l(q)$ takes $O(kn^{o(1)})$ time
    \item The \textsc{Retrieve} operation takes $O(|\mathcal{H}_l(q)|)$ time. By Lemma~\ref{lem:expect_number_points_level_set_formal}, $|\mathcal{H}_l(q)|=O(R)$
\end{itemize}

The running time of one iteration is $O(kn^{o(1)}+R)$

The final running of this procedure is $L\cdot O(kn^{o(1)}+R)=O(Lkn^{o(1)}+LR)$.

Thus, we complete the proof.
\end{proof}

Based on the running time of \textsc{Recover} in LSH above, we prove the running time of \textsc{Query} procedure in \textsc{DynamicKDE}. 

\begin{lemma}[Query part of Theorem~\ref{thm:main_result}, formal version of Lemma~\ref{lem:dynamic_KDE_query}]\label{lem:dynamic_KDE_query_formal}
Given a query $q\in\R^d$, the \textsc{Query} of the data-structure \textsc{DynamicKDE} (Algorithm~\ref{alg:dynamic_KDE_query}) runs in (expected) time 
\begin{align*}
    O(\epsilon^{-2}n^{o(1)}\log(1/f_{\mathsf{KDE}})\cdot f_{\mathsf{KDE}}^{-o(1)}\cdot\mathrm{cost}(K)\log^3 n).
\end{align*}
\end{lemma}

\begin{proof}[Proof]
First, the algorithm do a for loop with $K_1=O(\epsilon^{-2}\log n\cdot f_{\mathsf{KDE}}^{-o(1)})$ iterations. 

In each iteration, the running time consists of three parts. 

{\bf Part 1.} The running time from line~\ref{lin:second_loop} to line~\ref{lin:end_second_loop}, which is a for loop with $R$ iterations. In each iteration, the running time comes from
\begin{itemize}
    \item By Lemma~\ref{lem:recover_point_from_q}, we replace $L$ with $K_2,j=O(\mathrm{cost}(K)\log n)$, $J=O(\log(1/f_{\mathsf{KDE}}))$ and $k$ with $k_j=O(\log n)$.
    Thus  line~\ref{lin:evaluate_recover} takes $O(n^{o(1)}\mathrm{cost}(K)\log^2 n)$ time.
    \item line~\ref{lin:choose_L_j_point} takes $O(|\mathcal{H}_{a,r}.\mathcal{R}|)$ time. By Lemma~\ref{lem:upper_bound_recovered_point}, $|\mathcal{H}_{a,r}.\mathcal{R}|=O(1)$. Thus the running time of line~\ref{lin:choose_L_j_point} is $O(1)$.
\end{itemize}

Thus the running time of this for loop is $R \cdot O(n^{o(1)}\mathrm{cost}(K)\log^2 n) = O(n^{o(1)}\log(1/f_{\mathsf{KDE}})\cdot\mathrm{cost}(K)\log^2 n)$, where we use $R = O(\log(1/f_{\mathsf{KDE}}))$.

{\bf Part 2.} The running time from line~\ref{lin:third_loop} to line~\ref{lin:end_third_loop}, which is a forloop with $|\mathcal{S}|$ iterations. In each iteration, the running time is $O(1)$. By Lemma~\ref{lem:upper_bound_recovered_point}, $|\mathcal{S}|=O(R)$. Thus the running time of this forloop is $O(\log(1/f_{\mathsf{KDE}}))$, where we use $R = O(\log(1/f_{\mathsf{KDE}}))$.  

{\bf Part 3.} The running time of line~\ref{lin:recover_point_J+1},~\ref{lin:add_point_to_S} and~\ref{lin:output_Z_a} is $O(1)$

The final running time of \textsc{Query} is
\begin{align*}
    &~K_1\cdot ({\bf Part 1}+{\bf Part 2}+{\bf Part 3})\\
    =&~K_1\cdot O(n^{o(1)}\log(1/f_{\mathsf{KDE}})\cdot\mathrm{cost}(K)\log^2 n
    \\
    & ~ +O(\log(1/f_{\mathsf{KDE}}))+O(1))\\
    =&~O(\epsilon^{-2}n^{o(1)}\log(1/f_{\mathsf{KDE}})\cdot f_{\mathsf{KDE}}^{-o(1)}\cdot\mathrm{cost}(K)\log^3 n)
\end{align*}
where the first step follows directly from the running time of three parts, and the last step follows from $K_1=O(\epsilon^{-2}\log n\cdot f_{\mathsf{KDE}}^{-o(1)})$

Thus, we complete the proof.

\end{proof}

\subsection{LSH data structure}\label{app:data_structure_lsh}

% \Junze{This section is the same as Section 4.1.}

In this section, we present the LSH data structures with the following procedures:

{\bf Initialize} Given a data set $\{x_1, \cdots, x_n\} \in \R^d$ and integral parameters $k, L$, it first invokes private procedure \textsc{ChooseHashFunc}. The idea behind this is to amplify the "sensitivity" of hashing by concatenating $k$ basic hashing functions from the family $\mathcal{H}$(Algorithm~\ref{alg:LSH_public_app} line~\ref{lin:basic_hash_family}) into a new function. Thus we obtain a family of "augmented" hash function $\mathcal{H}_l, l \in [L]$ (Algorithm~\ref{alg:LSH_private_app} line~\ref{lin:LSH_sample_k_functions}). We follow by \textsc{ConstructHashTable} in which we hash each point $x_i$ using the hashing function $\mathcal{H}_{l}$. Then we obtain $L$ hash tables corresponding to $L$ hash functions which can be updated quickly.

{\bf Recover} Given a query $q \in \R^d$, it finds the bucket where $q$ is hashed by $\mathcal{H_l}$ and retrieve all the points in the bucket according to hashtable $\mathcal{T}_l$. This operation applies to all $L$ hashtables.

{\bf UpdateHashTable} Given a new data point $z \in \R^d$ and index $i \in [n]$, it repeats the following operations for all $l \in [L]$: find bucket $\mathcal{H}_l(z)$ and insert point $z$; find bucket $\mathcal{H}_l(x_i)$ and delete point $x_i$.

\begin{algorithm}[!ht]\caption{LSH, members and public procedures}\label{alg:LSH_public_app}
\begin{algorithmic}[1]
\State {\bf data structure} \textsc{LSH}
\State {\bf members}
    \State \hspace{4mm} $d,n \in \mathbb{N}_+$ \Comment{$d$ is dimension, $n$ is number of data points}
    \State \hspace{4mm} $K,L\in \mathbb{N}_+$ \Comment{$K$ is amplification factor, $L$ is number of repetition for hashing}
    \State \hspace{4mm} $p_{\mathrm{near}},p_{\mathrm{far}}\in (0,1)$ \Comment{Collision probability}
    \State \hspace{4mm} For $l \in L$, $\mathcal{T}_l:=[n]$ \Comment{Hashtable recording data points hashed by $\mathcal{H}_l$}
    \State \hspace{4mm} $\mathcal{R}:=[n]$ \Comment{retrieved points}
    \State \hspace{4mm} $\mathcal{H}:=\{f\in\mathcal{H}:\mathbb{R}^{d}\rightarrow[M]\}$ \Comment{$M$ is number of buckets for hashing family $\mathcal{H}$}
    \State \hspace{4mm} For $l \in [L]$, $\mathcal{H}_{l} \in \mathcal{H}^K$ \label{lin:basic_hash_family} \Comment{Family of amplified hash functions with at most $M^K$ non-empty buckets}
    \State \hspace{4mm} For $b \in [M^K]$, $\mathcal{S}_b:=$AVL tree \Comment{Use AVL tree to store points in bucket}
\State {\bf end members}

\State
\State {\bf public}
\Procedure{\textsc{Initialize}}{$\{x_i\}_{i\in[n]}\subset \mathbb{R}^d, k,L\in \mathbb{N}_+$}\label{lin:LSH_initialize}
 
    \State \textsc{ChooseHashFunc}($k,L$)\label{lin:LSH_intialize_choose_hash_func}
    \State \textsc{ConstructHashTable}($\{x_i\}_{i\in[n]}$)\label{lin:LSH_initialize_construct_hash_table}
\EndProcedure

\State
\Procedure{\textsc{Recover}}{$q\in\mathbb{R}^d$}\label{alg:LSH_recover}
    \State $\mathcal{R} \leftarrow 0$
    \For{$l\in[L]$}
        \State $\mathcal{R}\leftarrow \mathcal{R} \cup \mathcal{T}_{l}$.\textsc{Retrieve}($\mathcal{H}_{l}(q)$) \Comment{Find the bucket $\mathcal{H}_{l}(q)$ in $\mathcal{T}_l$ and retrieve all points}\label{lin:LSH_retrieve}
    \EndFor
\EndProcedure

\State
\Procedure{\textsc{UpdateHashTable}}{$z\in\mathbb{R}^d, i\in[n]$}\label{lin:update_hashtable}
    \For{$l\in [L]$}
        \State $\mathcal{H}_{l}(z)$.\textsc{Insert}($z$)\label{lin:insert} \Comment{$\mathcal{H}_{l}(z)$ denotes the bucket that $z$ is mapped to}
        \State $\mathcal{H}_{l}(x_i)$.\textsc{Delete}($x_i$)\label{lin:delete}
    \EndFor
\EndProcedure
\State {\bf end data structure}
\end{algorithmic}
\end{algorithm}

Next, we provide a private procedure of LSH in Algorithm~\ref{alg:LSH_private_app}.

\begin{algorithm}[!ht]\caption{LSH, private procedures}\label{alg:LSH_private_app}
\begin{algorithmic}[1]
\State {\bf data structure} \textsc{LSH}
\State
\State {\bf private}
\Procedure{\textsc{ChooseHashFunc}}{$k,L\in \mathbb{N}_+$}\label{lin:choose_hash_func}
    \For{$l \in [L]$}
        \State \Comment{Amplify hash functions by concatenating}
        \State $\mathcal{H}_{l} \leftarrow$ sample $k$ hash functions $(f_{1,l},f_{2,l},\cdots,f_{k,l})$ from $\mathcal{H}$ \label{lin:LSH_sample_k_functions}
    \EndFor
\EndProcedure

\State
\Procedure{\textsc{ConstructHashTable}}{$\{x_i\}_{i\in[n]}\subset \mathbb{R}^d$}\label{lin:construct_hash_table}
    \For{$l\in [L]$}
        \For{$i\in [n]$}
            \State $\mathcal{H}_l(x_i)$.\textsc{Insert}($x_i$) \label{lin:find_bucket_insert_element}
            \State $\mathcal{T}_l\leftarrow \mathcal{T}_l\cup \mathcal{H}_l(x_i)$ \Comment{Creat hashtable by aggregating buckets} \label{lin:aggregate_hash_table}
        \EndFor
    \EndFor
\EndProcedure

\State {\bf end data structure}
\end{algorithmic}
\end{algorithm}

%% file: correctness.tex
\section{Correctness: A Brief Summary}\label{sec:correctness}

In this section, we briefly discuss the correctness of the algorithm from data independence (Section~\ref{sec:correctness:init_update}), unbiasedness (Section~\ref{sec:correctness:unbiasedness_query}) and variance (Section~\ref{sec:correctness:variance_query}) aspects.

\subsection{Data Independence of \textsc{Initialize} and \textsc{Update}.}\label{sec:correctness:init_update}

We will show that procedure \textsc{Initialize} is independent of the input data. Hence, it is convenient for procedure \textsc{Update} to change only a small part of the stored data, without re-initializing the whole data structure.
For \textsc{Initialize}, it first samples data points by doing $n$ rounds of Bernoulli trials on each data point (Algorithm~\ref{alg:dynamic_KDE_initialize}, line~\ref{lin:sample}). Then it invokes \textsc{LSH.Initialize} to hash the sampled data points into some buckets, which will be stored into an LSH instance $\mathcal{H}_{a,r}$ (Algorithm~\ref{alg:dynamic_KDE_initialize}, Line~\ref{lin:instance_LSH}).
For \textsc{Update}, it finds the bucket where the old data point is hashed (Algorithm~\ref{alg:LSH_public_app}, Line~\ref{lin:update_hashtable} and  Line~\ref{lin:insert}) and replaces it with the new one. Thus procedure \textsc{Update} maintains the initialized structure.

\subsection{Unbiasedness of \textsc{Query}}\label{sec:correctness:unbiasedness_query}

We now present %prove
the unbiasedness of the estimator (up to some inverse polynomial error).

\begin{lemma}[unbiasedness of the \textsc{Query}, informal version of Lemma~\ref{lem:unbias_query_formal}]\label{lem:unbias_query}
Given $f_{\mathsf{KDE}}^{*} \in(0,1)$, $f_{\mathsf{KDE}} \geq f_{\mathsf{KDE}}^{*}$, $\epsilon \in(f_{\mathsf{KDE}}^{10}, 1)$ and $q \in \mathbb{R}^{d}$, we claim that estimator $T_{a}$ for any $a \in[K_{1}]$ constructed in line~\ref{lin:output_Z_a} Algorithm~\ref{alg:dynamic_KDE_query} satisfies $(1-n^{-9}) n f_{\mathsf{KDE}}^{*} \leq \E[T_{a}] \leq n f_{\mathsf{KDE}}^{*}$.
\end{lemma}

\subsection{Variance Bound for \textsc{Query}}\label{sec:correctness:variance_query}

We present the variance bound of our estimator.

\begin{lemma}[Variance bound for \textsc{Query}, informal version of Lemma~\ref{lem:variance_query_formal}]\label{lem:variance_query}
Given $f_{\mathsf{KDE}}^{*} \in(0,1)$, $f_{\mathsf{KDE}}/4 \leq f_{\mathsf{KDE}}^{*} \leq f_{\mathsf{KDE}}$, $\epsilon \in(f_{\mathsf{KDE}}^{10}, 1)$ and $q \in \mathbb{R}^{d}$, 
 
\textsc{Query} can output a $(1 \pm \epsilon)$-factor approximation to $f_{\mathsf{KDE}}^{*}$.
\end{lemma}

% For more details, we refer readers to \cite{ckns20} on why we can have such a quantity of $f_{\mathsf{KDE}}$.

%% file: app_correctness.tex
\section{Correctness: Details}\label{app:correctness}

The goal of this section is to prove the correctness of our algorithms. In Section~\ref{app:correctness:unbiasedness_query}, we provide the proof of unbiasedness for the query. In Section~\ref{app:correctness:variance_query}, we provide the proof of variance bound for the query.

\subsection{Unbiasedness of \textsc{Query}}\label{app:correctness:unbiasedness_query}

In this section, we prove that the estimator returned by \textsc{Query} is unbiased.

\begin{lemma}[unbiasedness of the \textsc{Query}, formal version of Lemma~\ref{lem:unbias_query}]\label{lem:unbias_query_formal}
For every $f_{\mathsf{KDE}}^{*} \in(0,1)$, every $f_{\mathsf{KDE}} \geq f_{\mathsf{KDE}}^{*}$, every $\epsilon \in(f_{\mathsf{KDE}}^{10}, 1)$, every $q \in \mathbb{R}^{d}$, estimator $T_{a} = \sum_{x_{i}\in\mathcal{S}}\frac{w_i}{p_i}$ for any $a \in[K_{1}]$ constructed in line~\ref{lin:output_Z_a} Algorithm~\ref{alg:dynamic_KDE_query} satisfies the following:

\begin{align*}
(1-n^{-9}) n f_{\mathsf{KDE}}^{*} \leq \E[T_{a}] \leq n f_{\mathsf{KDE}}^{*}
\end{align*}
\end{lemma}

\begin{proof}
Let $\mathcal{E}$ be the event that all the points are sampled. Let $T:=T_{a}$ (see line~\ref{lin:output_Z_a} Algorithm~\ref{alg:dynamic_KDE_query}). By Lemma~\ref{lem:lower_bound_recovered_point} and union bound, we have
\begin{align*}
  \Pr[\mathcal{E}] \geq 1-n^{-9} 
\end{align*}

Thus we obtain $\E[T]=\sum_{i=1}^{n} \frac{\E[\chi_{i}]}{p_{i}} w_{i}$ and $(1-n^{-9}) p_{i} \leq \E[\chi_{i}] \leq p_{i}$, where $\chi_{i}=1$ is defined to be the event that point $p_{i}$ gets sampled and recovered in the phase corresponding to its weight level, and $\chi_{i}=0$ is defined to the contrary. Thus
\begin{align*}
  (1-n^{-9}) n f_{\mathsf{KDE}}^{*} \leq \E[T] \leq n f_{\mathsf{KDE}}^{*} 
\end{align*}
\end{proof}

\subsection{Variance Bound for \textsc{Query}}\label{app:correctness:variance_query}

The goal of this section is to prove the variance bound of the estimator.

\begin{lemma}[Variance bound for \textsc{Query}, formal version of Lemma~\ref{lem:variance_query} 
]\label{lem:variance_query_formal}
For every $f_{\mathsf{KDE}}^{*} \in(0,1)$, every $\epsilon \in(f_{\mathsf{KDE}}^{10}, 1)$, every $q \in \mathbb{R}^{d}$, using estimators $T_{a} = \sum_{x_{i}\in\mathcal{S}}\frac{w_i}{p_i}$, for $a \in[K_{1}]$ constructed in line~\ref{lin:output_Z_a} Algorithm~\ref{alg:dynamic_KDE_query}, where $f_{\mathsf{KDE}} / 4 \leq f_{\mathsf{KDE}}^{*} \leq f_{\mathsf{KDE}}$, one can output a $(1 \pm \epsilon)$-factor approximation to $f_{\mathsf{KDE}}^{*}$.
\end{lemma}

\begin{proof}
First, we have $T \leq n^{2} f_{\mathsf{KDE}}^{*}$, where equality holds when all the points are sampled and recovered in the phase of their weight levels. By Lemma~\ref{lem:unbias_query}, we have
\begin{align*}
\E[T ~|~ \mathcal{E}] \cdot \Pr[\mathcal{E}]+n^{2} f_{\mathsf{KDE}}^{*}(1-\Pr[\mathcal{E}]) \geq \E[T]    
\end{align*}

Also, we have
\begin{align*}
\E[T ~|~ \mathcal{E}] \leq \frac{\E[T]}{\Pr[\mathcal{E}]} \leq \frac{n f_{\mathsf{KDE}}^{*}}{\Pr[\mathcal{E}]}=n f_{\mathsf{KDE}}^{*}(1+o(1 / n^{9}))
\end{align*}

Then, we have

\begin{align*}
\E[T^{2}] &=\E[(\sum_{p_{i} \in P} \chi_{i} \frac{w_{i}}{p_{i}})^{2}] \\
&=\sum_{i \neq j} \E[\chi_{i} \chi_{j} \frac{w_{i} w_{j}}{p_{i} p_{j}}]+\sum_{i \in[n]} \E[\chi_{i} \frac{w_{i}^{2}}{p_{i}^{2}}] \\
& \leq \sum_{i \neq j} w_{i} w_{j}+\sum_{i \in[n]} \frac{w_{i}^{2}}{p_{i}} \mathbb{I}[p_{i}=1]+\sum_{i \in[n]} \frac{w_{i}^{2}}{p_{i}} \mathbb{I}[p_{i} \neq 1] \\
& \leq(\sum_{i} w_{i})^{2}+\max _{i}\{\frac{w_{i}}{p_{i}} \mathbb{I}[p_{i} \neq 1]\} \sum_{i \in[n]} w_{i} \\
& \leq 2 n^{2}(f_{\mathsf{KDE}}^{*})^{2}+\max _{j \in[J], p_{i} \in L_{j}}\{w_{i} 2^{j+1}\} n f_{\mathsf{KDE}} \cdot n f_{\mathsf{KDE}}^{*} \\
& \leq 4 n^{2} f_{\mathsf{KDE}}^{2} \end{align*}
where the first step follows definition of $T$, the second step follows from expanding the square of summation, the third step follows from $\chi_{i} \chi_{j} \leq 1$, the fourth step follows from $\frac{w_{i}^{2}}{p_{i}} \mathbb{I}[p_{i}=1] \leq w_i^2$ and $(\sum_{i} w_{i})^{2} = \sum_{i \neq j} w_{i} w_{j} + \sum_{i \in[n]} {w_{i}^{2}}$, the fifth step follows from $n f_{\mathsf{KDE}}^{*} = (\sum_{i} w_{i})^{2} $ and $p_j \geq 1/(n \cdot 2^{j+1} f_{\mathsf{KDE}} )$ and $f_{\mathsf{KDE}} \geq f_{\mathsf{KDE}}^*$.  

\begin{align*}
    \E[Z^{2} ~|~ \mathcal{E}] \leq \frac{\E[Z^{2}]}{\Pr[\mathcal{E}]} \leq n^{2} f_{\mathsf{KDE}}^{2-o(1)}(1+o(1 / n^{9}))
\end{align*}

Now, we repeat this process for $K_{1}=\frac{C \log n}{\epsilon^{2}} \cdot f_{\mathsf{KDE}}^{-o(1)}$ times with constant $C$. Then, we have $(1 \pm \epsilon)$-factor approximation with higher success probability. We show that if we repeat the procedure $m$ times and take average, denoted as $\bar{T}$, we have:

\begin{align*}
& ~ \Pr[|\bar{T}-n f_{\mathsf{KDE}}^{*}| \geq \epsilon n f_{\mathsf{KDE}}^{*}] 
\\
\leq & ~ \Pr[|\bar{T}-\E[T]| \geq \epsilon n f_{\mathsf{KDE}}^{*}-|\E[T]-n f_{\mathsf{KDE}}^{*}|]
\\
\leq & ~ \Pr[|\bar{T}-\E[T]| \geq(\epsilon-n^{-9}) n f_{\mathsf{KDE}}^{*}] \\
\leq & ~ \frac{\E[\bar{T}^{2}]}{(\epsilon-n^{-9})^{2}(n^{2} f_{\mathsf{KDE}}^{*})^{2}} 
\\
\leq & ~ \frac{1}{m} \frac{64 n^{2}(f_{\mathsf{KDE}}^{*})^{2}}{(\epsilon-n^{-9})^{2}(n^{2} f_{\mathsf{KDE}}^{*})^{2}}
\end{align*}

where the first step follows from $ | \bar{T} - n f_{\mathsf{KDE}}^{*} | \leq |\bar{T} - \E[T]| + | \E[T] - n f_{\mathsf{KDE}}^{*} |$, the second step follows from $\E[T] \geq (1 - n^{-9} n f_{\mathsf{KDE}}^{*})$, the third step follows from Markov inequality and the last step follows from $\E[\bar{T}^2] \leq \E[T^2]/m \leq 4 n^2 f_{\mathsf{KDE}}^2 $ and $f_{\mathsf{KDE}} \leq 4 f_{\mathsf{KDE}}^*$.

We can repeat $m=O(\frac{1}{\epsilon^{2}})$ times to upper bound failure probability to $\delta$ and then take the median out of $O(\log (1 / \delta))$ means, where we assume $\delta = \frac{1}{\mathrm{poly}(n)}$.

\end{proof}

%% file: app_adversary.tex
\section{Adversary}\label{app:adversary}

In this section, we provide the detailed proofs for the lemmas in Section~\ref{sec:adversary}.

{\bf Starting Point} In Section~\ref{sec:correctness}, we have already obtained a query algorithm with constant success probability for a fixed query point.

\begin{lemma}[Starting with constant probability, restatement of Lemma~\ref{lem:single_estimator}]\label{lem:single_estimator_app}
Given $\epsilon \in (0,0.1)$, a query point $q \in \R^d$ and a set of data points $X = \{ x_{i}\}_{i=1}^{n} \subset \R^d$, let  $f_{\mathsf{KDE}}^*(q) := \frac{1}{|X|} \sum_{x \in X} f(x,q)$ be
an estimator $\mathcal{D}$ can answer the query which satisfies:
\begin{align*}
(1-\epsilon) \cdot f_{\mathsf{KDE}}^*(q) \leq \mathcal{D}.\textsc{query}(q, \epsilon) \leq (1 + \epsilon)\cdot f_{\mathsf{KDE}}^*(q)
\end{align*}
with probability $0.9$.
\begin{align*}
\end{align*}
\end{lemma}

\begin{proof}
By Lemma~\ref{lem:unbias_query_formal} and Lemma~\ref{lem:variance_query_formal}, our \textsc{Query} procedure can provide an estimator that answers kernel density estimation correctly with constant probability. 
\end{proof}

\paragraph{Boost the constant probability to high probability.}

Next, we begin to boost the success probability by repeating the query procedure and taking the median output.

\begin{lemma}[Boost the constant probability to high probability, restatement of Lemma~\ref{lem:fixed_points}]\label{lem:fixed_points_app}
Let $\delta_1 \in (0,0.1)$ denote the failure probability. Let $\epsilon \in (0,0.1)$ denote the accuracy parameter.
Given $L = O( \log(1/\delta_1) )$  estimators $\{\mathcal{D}_j\}_{j=1}^{L}$. For each fixed query point $q \in \R^d$, the median of queries from $L$ estimators satisfies that:
\begin{align*}
    (1-\epsilon) \cdot f_{\mathsf{KDE}}^*(q) \leq & ~  \mathrm{Median}(\{\mathcal{D}_j.\textsc{query}(q, \epsilon)\}_{j=1}^{L})
    \\
    \leq & ~ (1 + \epsilon)\cdot f_{\mathsf{KDE}}^*(q)
\end{align*}
with probability $1 - \delta_1$.
\end{lemma}

\begin{proof}
From Lemma~\ref{lem:single_estimator} we know each estimator $\mathcal{D}_j$ can answer the query that satisfies:
\begin{align*}
(1-\epsilon) \cdot f_{\mathsf{KDE}}^*(q) \leq \mathcal{D}.\textsc{query}(q, \epsilon) \leq (1 + \epsilon)\cdot f_{\mathsf{KDE}}^*(q)
\end{align*}
with probability $0.9$.

From the chernoff bound we know the median of $L =O( \log(1/\delta_1))$ queries from $\{\mathcal{D}_j\}_{j=1}^{L}$ satisfies:
\begin{align*}
    (1-\epsilon) \cdot f_{\mathsf{KDE}}^*(q) \leq & ~ \mathrm{Median}(\{\mathcal{D}_j.\textsc{query}(q, \epsilon)\}_{j=1}^{L}) 
    \\
    \leq & ~ (1 + \epsilon)\cdot f_{\mathsf{KDE}}^*(q)
\end{align*}
with probability $1 - \delta_1$.

Therefore, we complete the proof.
\end{proof}

\paragraph{From each fixed point to all the net points.}

So far, the success probability of our algorithm is still for a fix point. We will introduce $\epsilon$-net on a unit ball and show the high success probability for all the net points. 

\begin{fact}\label{fac:number_of_net_points_app}
Let $N$ denote the $\epsilon_0$-net of $\{ x \in \R^d ~|~ \| x \|_2 \leq 1 \}$. We use $|N|$ to denote the number of points in $N$. Then $|N|\leq (10/\epsilon_0)^d$.
\end{fact}

This fact shows that we can bound the size of an $\epsilon$-net with an inverse of $\epsilon$. We use this fact to conclude the number of repetitions we need to obtain the correctness of \textsc{Query} on all net points.

\begin{lemma}[From each fixed points to all the net points, restatement of Lemma~\ref{lem:net_points}]\label{lem:net_points_app}
Let $N$ denote the $\epsilon_0$-net of $\{ x \in \R^d ~|~ \| x \|_2 \leq 1 \}$. We use $|N|$ to denote the number of points in $N$. Given $L = \log(|N|/\delta)$  estimators $\{\mathcal{D}_j\}_{j=1}^{L}$. 

With probability $1 - \delta$, we have: for all $q \in N$, the median of queries from $L$ estimators satisfies that:
\begin{align*}
   (1-\epsilon) \cdot f_{\mathsf{KDE}}^*(q) \leq & ~ \mathrm{Median}(\{\mathcal{D}_j.\textsc{query}(q, \epsilon)\}_{j=1}^{L})
   \\
   & ~ \leq (1 + \epsilon)\cdot f_{\mathsf{KDE}}^*(q).
\end{align*}
\end{lemma}

\begin{proof}
There are $|N|$ points on the $d$ dimension $\epsilon$-net when $\| q\|_2 \leq 1$. From Lemma~\ref{lem:fixed_points_app} we know that for each query point $q$ on $N$, we have :
\begin{align*}
    (1-\epsilon) \cdot f_{\mathsf{KDE}}^*(q) \leq & ~ \mathrm{Median}(\{\mathcal{D}_j.\textsc{query}(q, \epsilon)\}_{j=1}^{L}) 
    \\
    \leq & ~ (1 + \epsilon)\cdot f_{\mathsf{KDE}}^*(q)
\end{align*}
with probability $1 - \delta/|N|$.

By union bound all $|N|$ points on $N$, we have:
\begin{align*}
    \forall \| q\|_2 \leq 1 : (1-\epsilon) \cdot f_{\mathsf{KDE}}^*(q) \leq & ~ \mathrm{Median}(\{\mathcal{D}_j.\textsc{query}(q, \epsilon)\}_{j=1}^{L}) 
    \\
    \leq & ~ (1 + \epsilon)\cdot f_{\mathsf{KDE}}^*(q)
\end{align*}
with probability $1 - \delta$.
\end{proof}

\paragraph{From net points to all points.}
With Lemma~\ref{lem:net_points_app}, we are ready to extend the correctness for net points to the whole unit ball. We demonstrate that all query points  
$\| q \|_2 \leq 1$ can be answered approximately with high probability in the following lemma.

\begin{lemma}[From net points to all points, restatement of Lemma \ref{lem:from_net_points_to_all_points}
]
Let $\epsilon \in (0,0.1)$. Let ${\cal L} \geq 1$. Let $\delta \in (0,0.1)$. Let $\tau \in [0,1]$. 
Given $L = O(\log(( \mathcal{L}/\epsilon \tau )^d/\delta))$  estimators $\{\mathcal{D}_j\}_{j=1}^{L}$, with probability $1 - \delta$, for all query points $\|p\|_2 \leq 1$,  we have the median of queries from $L$ estimators satisfies that:
\begin{align*}
    \forall \| p\|_2 \leq 1: \\
    & ~ (1-\epsilon) \cdot f_{\mathsf{KDE}}^*(p) \\
    \leq & ~ \mathrm{Median}(\{\mathcal{D}_j.\textsc{query}(q, \epsilon)\}_{j=1}^{L})
    \\
    \leq & ~ (1 + \epsilon)\cdot  f_{\mathsf{KDE}}^*(p).
\end{align*}
where $q$ is the closest net point of $p$.

\end{lemma}

\begin{proof}

We define an event $\xi$ to be the following,
\begin{align*}
    \forall q \in N, ~~~ (1-\epsilon) \cdot f_{\mathsf{KDE}}^*(q) \leq & ~ \mathrm{Median}(\{\mathcal{D}_j.\textsc{query}(q, \epsilon)\}_{j=1}^{L}) 
    \\
    \leq & ~ (1 + \epsilon)\cdot f_{\mathsf{KDE}}^*(q)
\end{align*}

Using Lemma~\ref{lem:net_points_app} with $L= \log(|N|/\delta)$, we know that 
\begin{align*}
    \Pr[~\text{~event~} \xi \text{~holds~} ] \geq 1 - \delta
\end{align*}

Using Fact~\ref{fac:number_of_net_points_app}, we know that
\begin{align*}
    L
    = & ~ \log(|N|/\delta ) \\
    = & ~ \log( ( 10 /\epsilon_0 )^d / \delta )\\ 
    = & ~ \log((10 \mathcal{L}/\epsilon \tau )^d/\delta)
\end{align*}
where the last step follows from $\epsilon_0 = \epsilon \tau / \mathcal{L}$. 

We condition the above event $E$ to be held. (Then the remaining proof does not depend on any randomness, for each and for all becomes the same.)

For each point $p \notin N$, there exists a $q \in N$ such that
\begin{align}\label{eq:diff_p_q_app}
    \| p - q \|_2 \leq \epsilon_0
\end{align}
For each $p$, we know
\begin{align*}
    & ~ |\mathrm{Median}_j\mathcal{D}_j.\textsc{query}(q, \epsilon) - f_{\mathsf{KDE}}^*(p) | \\
    \leq & ~ |\mathrm{Median}_j\mathcal{D}_j.\textsc{query}(q, \epsilon)-f_{\mathsf{KDE}}^*(q)|+|f_{\mathsf{KDE}}^*(q)-f_{\mathsf{KDE}}^*(p)|\\
    \leq & ~  \epsilon f_{\mathsf{KDE}}^*(q)+\mathcal{L}\cdot \| p - q \|_2 \\
    \leq & ~ \epsilon(f_{\mathsf{KDE}}^*(p)+\mathcal{L}\epsilon_0) + \mathcal{L} \cdot \epsilon_0 \\
    \leq & ~ \epsilon (f_{\mathsf{KDE}}^*(p)+2\tau) 
\end{align*}
where the first step follows from Lipschitz, the second step follows from Eq.~\eqref{eq:diff_p_q_app}, the third step follows from $\epsilon_0 \leq \epsilon \tau / \mathcal{L}$.

Using $ \forall j \in [L]: \mathcal{D}_j.\textsc{query}(p, \epsilon) \geq \tau$, we have 
\begin{align*}
    (1-3\epsilon) \cdot f_{\mathsf{KDE}}^*(p)  \leq & ~ \mathrm{Median}(\{\mathcal{D}_j.\textsc{query}(q, \epsilon)\}_{j=1}^{L})
    \\
    \leq & ~ (1 +3 \epsilon)\cdot f_{\mathsf{KDE}}^*(p) .
\end{align*}
Rescaling the $\epsilon$ completes the proof.
\end{proof}

Thus, we obtain an algorithm that could respond to adversary queries robustly.

%% file: app_lipschitz.tex
\section{Lipschitz}\label{app:lipschitz}

The goal of this section is to prove the Lipschitz property of the KDE function.

\subsection{Lipschitz property of KDE function}

\begin{lemma}
Suppose kernel function $f^*_{\mathsf{KDE}}:\R^d\times\R^d\rightarrow[0,1]$ satisfies the following properties:
\begin{itemize}
    \item Radial: there exists a funtion $f:\R\rightarrow[0,1]$ such that $f(p,q)=f(\|p-q\|_2)$, for all $p,q\in\R$.
    \item Decreasing: $f$ is decreasing
    \item Lipschitz: $f$ is $\mathcal{L}$-Lipschitz
\end{itemize}
Then KDE function $f_{\mathsf{KDE}}^*:\R^d\rightarrow [0,1]$, $f_{\mathsf{KDE}}^*(q):=\frac{1}{|P|}\sum_{p\in P}f(p,q)$ is $\mathcal{L}$-Lipschitz, i.e.
\begin{align*}
    |f_{\mathsf{KDE}}^*(q)-f_{\mathsf{KDE}}^*(q')|\leq \mathcal{L}\cdot\|q-q'\|_2
\end{align*}
\end{lemma}

\begin{proof}

For any $q,q' \in\R^d$, we have:

\begin{align*}
    &~|f_{\mathsf{KDE}}^*(q)-f_{\mathsf{KDE}}^*(q')|\\
    =&~|\frac{1}{|P|}\sum_{p\in P}f(p,q)-\frac{1}{|P|}\sum_{p\in P}f(p,q')|\\
    \leq&~\frac{1}{|P|}\sum_{p\in P}|f(p,q)-f(p,q')|\\
    =&~\frac{1}{|P|}\sum_{p\in P}|f(\|p-q\|_2)-f(\|p-q'\|_2)|\\
    \leq&~\frac{1}{|P|}\sum_{p\in P}\mathcal{L}\cdot|~\|p-q\|_2-\|p-q'\|_2 ~|\\
    \leq&~\frac{1}{|P|}\sum_{p\in P}\mathcal{L}\cdot\|q-q'\|_2\\
    =&~\mathcal{L}\cdot\|q-q'\|_2
\end{align*}

where the first step follows from the definition of $f_{\mathsf{KDE}}^*$, the second step follows from triangular inequality of absolute value, the third step follows from the property of radial kernel, the fourth step follows from the Lipschitz property of $f$, the fifth step follows from the triangular property of $L_2$ norm, and the last step follows from canceling $|P|$.

Thus, we complete the proof.
\end{proof}

%% file: main.bbl
\newcommand{\etalchar}[1]{$^{#1}$}
\begin{thebibliography}{WXW{\etalchar{+}}22}

\bibitem[ACSS20]{acss20}
Josh Alman, Timothy Chu, Aaron Schild, and Zhao Song.
\newblock Algorithms and hardness for linear algebra on geometric graphs.
\newblock In {\em 2020 IEEE 61st Annual Symposium on Foundations of Computer
  Science (FOCS)}, pages 541--552. IEEE, 2020.

\bibitem[AI06]{ai06}
Alexandr Andoni and Piotr Indyk.
\newblock Near-optimal hashing algorithms for approximate nearest neighbor in
  high dimensions.
\newblock In {\em 2006 47th annual IEEE symposium on foundations of computer
  science (FOCS'06)}, pages 459--468. IEEE, 2006.

\bibitem[AS23]{as23}
Josh Alman and Zhao Song.
\newblock Fast attention requires bounded entries.
\newblock {\em arXiv preprint arXiv:2302.13214}, 2023.

\bibitem[BIW19]{biw19}
Arturs Backurs, Piotr Indyk, and Tal Wagner.
\newblock Space and time efficient kernel density estimation in high
  dimensions.
\newblock {\em Advances in Neural Information Processing Systems}, 32, 2019.

\bibitem[CGC{\etalchar{+}}22]{cgc+18}
Benjamin Coleman, Benito Geordie, Li~Chou, RA~Leo Elworth, Todd Treangen, and
  Anshumali Shrivastava.
\newblock One-pass diversified sampling with application to terabyte-scale
  genomic sequence streams.
\newblock In {\em International Conference on Machine Learning}, pages
  4202--4218. PMLR, 2022.

\bibitem[CHM12]{chm12}
Yuan Cao, Haibo He, and Hong Man.
\newblock Somke: Kernel density estimation over data streams by sequences of
  self-organizing maps.
\newblock {\em IEEE transactions on neural networks and learning systems},
  23(8):1254--1268, 2012.

\bibitem[CHS20]{chs20}
Jaewoong Cho, Gyeongjo Hwang, and Changho Suh.
\newblock A fair classifier using kernel density estimation.
\newblock {\em Advances in Neural Information Processing Systems},
  33:15088--15099, 2020.

\bibitem[CIU{\etalchar{+}}21]{ciu21}
Tsz~Nam Chan, Pak~Lon Ip, Leong~Hou U, Byron Choi, and Jianliang Xu.
\newblock Sws: a complexity-optimized solution for spatial-temporal kernel
  density visualization.
\newblock {\em Proceedings of the VLDB Endowment}, 15(4):814--827, 2021.

\bibitem[CKNS20]{ckns20}
Moses Charikar, Michael Kapralov, Navid Nouri, and Paris Siminelakis.
\newblock Kernel density estimation through density constrained near neighbor
  search.
\newblock In {\em 2020 IEEE 61st Annual Symposium on Foundations of Computer
  Science (FOCS)}, pages 172--183. IEEE, 2020.

\bibitem[CLP{\etalchar{+}}20]{clp+20}
Beidi Chen, Zichang Liu, Binghui Peng, Zhaozhuo Xu, Jonathan~Lingjie Li, Tri
  Dao, Zhao Song, Anshumali Shrivastava, and Christopher Re.
\newblock Mongoose: A learnable lsh framework for efficient neural network
  training.
\newblock In {\em International Conference on Learning Representations}, 2020.

\bibitem[CMF{\etalchar{+}}20]{cmf+20}
Beidi Chen, Tharun Medini, James Farwell, Charlie Tai, Anshumali Shrivastava,
  et~al.
\newblock Slide: In defense of smart algorithms over hardware acceleration for
  large-scale deep learning systems.
\newblock {\em Proceedings of Machine Learning and Systems}, 2:291--306, 2020.

\bibitem[Cra01]{c01}
Kyle Cranmer.
\newblock Kernel estimation in high-energy physics.
\newblock {\em Computer Physics Communications}, 136(3):198--207, 2001.

\bibitem[CRV{\etalchar{+}}18]{crv+18}
Danielle~L Cantrell, Erin~E Rees, Raphael Vanderstichel, Jon Grant, Ram{\'o}n
  Filgueira, and Crawford~W Revie.
\newblock The use of kernel density estimation with a bio-physical model
  provides a method to quantify connectivity among salmon farms: spatial
  planning and management with epidemiological relevance.
\newblock {\em Frontiers in Veterinary Science}, page 269, 2018.

\bibitem[CS16]{cs16}
Efren~Cruz Cortes and Clayton Scott.
\newblock Sparse approximation of a kernel mean.
\newblock {\em IEEE Transactions on Signal Processing}, 65(5):1310--1323, 2016.

\bibitem[CS17]{cs17}
Moses Charikar and Paris Siminelakis.
\newblock Hashing-based-estimators for kernel density in high dimensions.
\newblock In {\em 2017 IEEE 58th Annual Symposium on Foundations of Computer
  Science (FOCS)}, pages 1032--1043. IEEE, 2017.

\bibitem[CS20]{cs20}
Benjamin Coleman and Anshumali Shrivastava.
\newblock Sub-linear race sketches for approximate kernel density estimation on
  streaming data.
\newblock In {\em Proceedings of The Web Conference 2020}, pages 1739--1749,
  2020.

\bibitem[CWS10]{cws10}
Yutian Chen, Max Welling, and Alex Smola.
\newblock Super-samples from kernel herding.
\newblock In {\em Proceedings of the Twenty-Sixth Conference on Uncertainty in
  Artificial Intelligence}, pages 109--116, 2010.

\bibitem[DJS{\etalchar{+}}22]{djs+22}
Yichuan Deng, Wenyu Jin, Zhao Song, Xiaorui Sun, and Omri Weinstein.
\newblock Dynamic kernel sparsifiers.
\newblock {\em arXiv preprint arXiv:2211.14825}, 2022.

\bibitem[DMS23]{dms23}
Yichuan Deng, Sridhar Mahadevan, and Zhao Song.
\newblock Randomized and deterministic attention sparsification algorithms for
  over-parameterized feature dimension.
\newblock {\em arXiv preprint arXiv:2304.04397}, 2023.

\bibitem[FC17]{fc17}
Christen~H Fleming and Justin~M Calabrese.
\newblock A new kernel density estimator for accurate home-range and
  species-range area estimation.
\newblock {\em Methods in Ecology and Evolution}, 8(5):571--579, 2017.

\bibitem[HIK{\etalchar{+}}21]{hallin2021classifying}
Anna Hallin, Joshua Isaacson, Gregor Kasieczka, Claudius Krause, Benjamin
  Nachman, Tobias Quadfasel, Matthias Schlaffer, David Shih, and Manuel
  Sommerhalder.
\newblock Classifying anomalies through outer density estimation (cathode).
\newblock {\em arXiv preprint arXiv:2109.00546}, 2021.

\bibitem[HL18]{hl18}
Yaoyao He and Haiyan Li.
\newblock Probability density forecasting of wind power using quantile
  regression neural network and kernel density estimation.
\newblock {\em Energy conversion and management}, 164:374--384, 2018.

\bibitem[HS08]{hs08}
Christoph Heinz and Bernhard Seeger.
\newblock Cluster kernels: Resource-aware kernel density estimators over
  streaming data.
\newblock {\em IEEE Transactions on Knowledge and Data Engineering},
  20(7):880--893, 2008.

\bibitem[IM98]{im98}
Piotr Indyk and Rajeev Motwani.
\newblock Approximate nearest neighbors: towards removing the curse of
  dimensionality.
\newblock In {\em Proceedings of the thirtieth annual ACM symposium on Theory
  of computing}, pages 604--613, 1998.

\bibitem[KAP22]{kap22}
Matti Karppa, Martin Aum{\"u}ller, and Rasmus Pagh.
\newblock Deann: Speeding up kernel-density estimation using approximate
  nearest neighbor search.
\newblock In {\em International Conference on Artificial Intelligence and
  Statistics}, pages 3108--3137. PMLR, 2022.

\bibitem[LWZ{\etalchar{+}}23]{lwz+23}
Zirui Liu, Guanchu Wang, Shaochen Zhong, Zhaozhuo Xu, Daochen Zha, Ruixiang
  Tang, Zhimeng Jiang, Kaixiong Zhou, Vipin Chaudhary, Shuai Xu, et~al.
\newblock Winner-take-all column row sampling for memory efficient adaptation
  of language model.
\newblock {\em arXiv preprint arXiv:2305.15265}, 2023.

\bibitem[MFS{\etalchar{+}}17]{mf+17}
Krikamol Muandet, Kenji Fukumizu, Bharath Sriperumbudur, Bernhard
  Sch{\"o}lkopf, et~al.
\newblock Kernel mean embedding of distributions: A review and beyond.
\newblock {\em Foundations and Trends{\textregistered} in Machine Learning},
  10(1-2):1--141, 2017.

\bibitem[MRL95]{mrl95}
Young-Il Moon, Balaji Rajagopalan, and Upmanu Lall.
\newblock Estimation of mutual information using kernel density estimators.
\newblock {\em Physical Review E}, 52(3):2318, 1995.

\bibitem[PT20]{pt20}
Jeff~M Phillips and Wai~Ming Tai.
\newblock Near-optimal coresets of kernel density estimates.
\newblock {\em Discrete \& Computational Geometry}, 63(4):867--887, 2020.

\bibitem[SRB{\etalchar{+}}19]{srb+19}
Paris Siminelakis, Kexin Rong, Peter Bailis, Moses Charikar, and Philip Levis.
\newblock Rehashing kernel evaluation in high dimensions.
\newblock In {\em International Conference on Machine Learning}, pages
  5789--5798. PMLR, 2019.

\bibitem[SS21]{ss21}
Ryan Spring and Anshumali Shrivastava.
\newblock Mutual information estimation using lsh sampling.
\newblock In {\em Proceedings of the Twenty-Ninth International Conference on
  International Joint Conferences on Artificial Intelligence}, pages
  2807--2815, 2021.

\bibitem[SSX21]{ssx21}
Anshumali Shrivastava, Zhao Song, and Zhaozhuo Xu.
\newblock Sublinear least-squares value iteration via locality sensitive
  hashing.
\newblock {\em arXiv preprint arXiv:2105.08285}, 2021.

\bibitem[SXYZ22]{sxyz22}
Zhao Song, Zhaozhuo Xu, Yuanyuan Yang, and Lichen Zhang.
\newblock Accelerating frank-wolfe algorithm using low-dimensional and adaptive
  data structures.
\newblock {\em arXiv preprint arXiv:2207.09002}, 2022.

\bibitem[SXZ22]{sxz22}
Zhao Song, Zhaozhuo Xu, and Lichen Zhang.
\newblock Speeding up sparsification using inner product search data
  structures.
\newblock {\em arXiv preprint arXiv:2204.03209}, 2022.

\bibitem[vdBSZ23]{bsz23}
Jan van~den Brand, Zhao Song, and Tianyi Zhou.
\newblock Algorithm and hardness for dynamic attention maintenance in large
  language models.
\newblock {\em arXiv e-prints}, pages arXiv--2304, 2023.

\bibitem[WHM{\etalchar{+}}09]{whm+09}
Meng Wang, Xian-Sheng Hua, Tao Mei, Richang Hong, Guojun Qi, Yan Song, and
  Li-Rong Dai.
\newblock Semi-supervised kernel density estimation for video annotation.
\newblock {\em Computer Vision and Image Understanding}, 113(3):384--396, 2009.

\bibitem[WLG19]{wlg19}
Qin Wang, Wen Li, and Luc~Van Gool.
\newblock Semi-supervised learning by augmented distribution alignment.
\newblock In {\em Proceedings of the IEEE/CVF international conference on
  computer vision}, pages 1466--1475, 2019.

\bibitem[WXW{\etalchar{+}}22]{wxw+22}
Zhuang Wang, Zhaozhuo Xu, Xinyu Wu, Anshumali Shrivastava, and TS~Eugene Ng.
\newblock Dragonn: Distributed randomized approximate gradients of neural
  networks.
\newblock In {\em International Conference on Machine Learning}, pages
  23274--23291. PMLR, 2022.

\bibitem[XCL{\etalchar{+}}21]{xcl+21}
Zhaozhuo Xu, Beidi Chen, Chaojian Li, Weiyang Liu, Le~Song, Yingyan Lin, and
  Anshumali Shrivastava.
\newblock Locality sensitive teaching.
\newblock {\em Advances in Neural Information Processing Systems}, 34, 2021.

\bibitem[XSS21]{xss21}
Zhaozhuo Xu, Zhao Song, and Anshumali Shrivastava.
\newblock Breaking the linear iteration cost barrier for some well-known
  conditional gradient methods using maxip data-structures.
\newblock {\em Advances in Neural Information Processing Systems}, 34, 2021.

\bibitem[Zha22]{z22}
Lichen Zhang.
\newblock Speeding up optimizations via data structures: Faster search, sample
  and maintenance.
\newblock Master's thesis, Carnegie Mellon University, 2022.

\end{thebibliography}
